%% file: main.tex
\documentclass[letterpaper,11pt]{article}
\usepackage{times}
\usepackage{amssymb,amsbsy,amsfonts,amsmath,amsthm,xspace}
\usepackage{mathrsfs}
\usepackage{graphicx}
\usepackage{caption}
\usepackage{setspace}
\usepackage[boxed]{algorithm2e}
\usepackage{enumitem}
\usepackage{pifont}

\usepackage{subcaption}
\usepackage{multirow}
\usepackage{epstopdf}
\usepackage{epsfig}
\usepackage{url}
\usepackage[toc,page]{appendix}
\usepackage{float}
\usepackage{natbib}
\usepackage{color}
\usepackage{verbatim}
\usepackage{authblk}
\usepackage[normalem]{ulem}

\theoremstyle{definition}
\newtheorem{thm}{Theorem}

\newtheorem{prop}{Proposition}
\newtheorem{coro}{Corollary}

\newtheorem{rem}{Remark}
\newtheorem{ex}{Example}

\newcommand{\e}{\mathbb{E}}
\newcommand{\p}{\mathbb{P}}

\newcommand{\ccal}{\mathcal{C}}
\newcommand{\lcal}{\mathcal{L}}

\newcommand{\mcal}{\mathcal{M}}

\newcommand{\tcal}{\mathcal{T}}

\newcommand{\bR}{\mathbb{R}}

\newcommand{\diag}{\text{diag}}

\DeclareMathOperator*{\argmin}{argmin}

\newcommand{\norm}[1]{\Vert{#1}\Vert}

\newcommand{\be}{\begin{equation}}
\newcommand{\ee}{\end{equation}}
\newcommand{\bes}{\begin{equation*}}
\newcommand{\ees}{\end{equation*}}
\newcommand{\ba}{\begin{align}}
\newcommand{\ea}{\end{align}}
\newcommand{\bas}{\begin{align*}}
\newcommand{\eas}{\end{align*}}

\def\text#1{\mbox{\rm #1}}

\newcommand{\printfnsymbol}[1]{%
  \textsuperscript{\@fnsymbol{#1}}%
}

\usepackage{hyperref}
\hypersetup{
    colorlinks=true,
    linkcolor=blue,
    filecolor=magenta,      
    urlcolor=cyan,
    citecolor=blue
}
 
\title{\textbf{Network Estimation by Mixing: Adaptivity and More}
\thanks{Both authors contributed equally to this work.}}

\author{Tianxi Li\\ Department of Statistics, University of Virginia
\and Can M. Le\\ Department of Statistics, University of California, Davis 
}
\date{\today}

\begin{document}

\maketitle

\begin{abstract}
\noindent
Networks analysis has been commonly used to study the interactions between units of complex systems. One problem of particular interest is learning the network's underlying connection pattern given a single and noisy instantiation. While many methods have been proposed to address this problem in recent years, they usually assume that the true model belongs to a known class, which is not verifiable in most real-world applications. Consequently, network modeling based on these methods either suffers from model misspecification or relies on additional model selection procedures that are not well understood in theory and can potentially be unstable in practice. To address this difficulty, we propose a mixing strategy that leverages available arbitrary models to improve their individual performances. The proposed method is computationally efficient and almost tuning-free; thus, it can be used as an off-the-shelf method for network modeling. We show that the proposed method performs equally well as the oracle estimate when the true model is included as individual candidates. More importantly, the method remains robust and outperforms all current estimates even when the models are misspecified. Extensive simulation examples are used to verify the advantage of the proposed mixing method. Evaluation of link prediction performance on 385 real-world networks from six domains also demonstrates the universal competitiveness of the mixing method across multiple domains.
\end{abstract}

\section{Introduction}
\input{Intro}

\section{The network mixing method and its properties}\label{sec:method}
\input{Setup}

\section{Simulation examples}\label{sec:simulation}

\input{Simulation}

\section{Link prediction in real-world networks}\label{sec:data}
\input{data}

\section{Discussion}\label{sec:disc}
\input{discussion}

\section*{Acknowledgement}
C. M. Le is supported in part by the NSF grant DMS-2015134. T. Li is supported in part by the NSF grant DMS-2015298 and the Quantitative Collaborative Award from the College of Arts and Sciences at the University of Virginia.

\bibliography{CommonBib}{}
\bibliographystyle{abbrvnat}

\begin{appendix}
\include{Proof}
\end{appendix}

\end{document}

%% file: Intro.tex
Networks are widely used to represent complex interactions between entities, and network analysis has been an intensively studied topic in statistics, computer science, physics, social science, and many areas in recent years. Analyzing the structures of network data can render salient insights about relation formulation or interaction mechanisms between individuals. Network analysis has been used in various applications in science, engineering, and social studies, revealing many new aspects in these studies  \citep{barabasi1999emergence,steglich2006applying,newman2010networks,fortunato2010community,lu2011link}.

Due to the noisy nature of network data, statistical modeling has been a common approach network analysis. Multiple probabilistic frameworks have been proposed for network data (e.g., \cite{barabasi1999emergence,Bollobas2007,crane2018probabilistic,ghafouri2020survey}). These frameworks are understood as  approximations of real-world network formulation from different aspects with pros and cons in various situations. Our discussion in this paper is embedded in arguably  one of the most popular frameworks for statistical network modeling --  the so-called ``inhomogeneous Erd\H{o}s-R\'{e}nyi model" \citep{Bollobas2007,goldenberg2010survey,newman2010networks}. This framework includes as special cases several important models in the network literature, such as the stochastic block model (SBM) \citep{holland1983stochastic} and its variants \citep{airoldi2008mixed,karrer2011stochastic,jin2017estimating,sengupta2018block,noroozi2019sparse} for community structures,  the random dot product model \citep{young2007random} and latent space model \citep{hoff2002latent, hoff2008modeling} for latent vector spaces, and the graphon model node-exchangeable graphs \citep{aldous1981representations,Lovasz&Szegedy.2006, diaconis2007graph}.

%Assuming that the network is generated from the inhomogeneous Erd\H{o}s-R\'{e}nyi model or the graphon model, a large body of methods have been developed to estimate the mean matrix $P=\mathbb{E} A$ or the link function $g$ from the adjacency matrix \citep{chatterjee2015matrix,airoldi2013stochastic,wolfe2013nonparametric,zhang2015estimating,gao2015rate,ma2020universal}.

While effective modeling strategies have been proposed and studied in many settings, it is observed that  real-world networks may exhibit a much wider range of patterns \citep{ugander2013subgraph,ghasemian2020stacking,miao2021informative}  and it is generally difficult to know which model is the proper one, or whether there is a proper choice. Therefore, an indispensable task in network modeling is to select a proper model or algorithm to use.  Depending on the scope of generality, model selection procedures have been proposed in different categories. The model-based methods are designed under a family of models, such as testing procedures \citep{bickel2013hypothesis, gao2017testing,lei2016goodness, mukherjee2017testing,Banerjee&Ma.optimal.test.2017, jin2019optimal,Amini.chisquare2020} and criterion-based methods \citep{saldana2014how,wang2015likelihood,le2015estimating,yan2017provable}. The more general model selection approach is cross-validation \citep{chen2014network, li2016network, chang2020discussion}, which can be used to compare models across different families. Cross-validation intuitively seeks the most proper model for the data even if none of the models under consideration are true. Despite the many good properties derived from these methods,  model selection can be unstable \citep{breiman1996bagging} in the sense that  a small perturbation of the data can result in selecting a different model. Moreover, even if one does know the correct class of candidate models, it is not always possible to effectively fit such models due to various constraints. For example, when the network is sparse, it is known that one cannot accurately estimate an SBM with a large number of communities \citep{rohe2011spectral,lei2014consistency, gao2015achieving,li2018hierarchical} or a general graphon structure \citep{chatterjee2015matrix,zhang2015estimating,gao2015rate}. 

To address the aforementioned shortcomings, this paper aims for an ``off-the-shelf" and principled network modeling strategy. From a practical perspective, it is computationally efficient for large-scale networks and does not require careful tuning or model specification. From the theoretical perspective, our strategy can be general enough to incorporate the well-studied models in the literature with a theoretical guarantee on performance. These properties are achieved by a simple but powerful idea: instead of selecting a single model, we take a constrained linear combination of all candidate models in a data-driven way, such that the aggregated estimate is provably more stable and adaptive to different underlying models.

Model aggregation is not unique to network data. The idea has been studied in statistical learning on metric space data, such as density estimation and regression problems, often known as the ``mixing" or ``stacking" procedure \citep{stone1974cross,wolpert1992stacked,breiman1996stacked,leblanc1996combining, catoni1997mixture,haussler1998sequential,yang2000mixing,yang2001adaptive,juditsky2000functional, Tsybakov2003,bunea2007aggregation}. 
However, the mixing strategy in network settings has not been well studied, except for the recent empirical work of \cite{ghasemian2020stacking}. Designing a network version of mixing estimation as an ``off-the-shelf" method is non-trivial. The previous mixing strategies either fail to produce good theoretical guarantees in network settings \citep{Tsybakov2003} or are computationally prohibitive  \citep{yang2000mixing,yang2001adaptive,Tsybakov2003,bunea2007aggregation,vanderLaan.superlearner.2007} and require additional tuning \citep{bunea2007aggregation}. The unique nature of network data (e.g., no i.i.d sample, discreteness, sparsity) also require special methodological designs and a different approach to theoretical analysis.

Our mixing methods rely on splitting the network node-pairs into a training and a validation data set, as in \cite{li2016network}, although other methods may be considered \citep{Spielman&Shrivastava2011,Rohe&Qin2013,Le.edge.sampling2017}. Multiple individual models are fit using the training data and then aggregated according to the validation data. A simple aggregation approach with exponential weights is introduced to achieve the basic model adaptivity, an ideal property one could hope for the cross-validation procedure. Then, based on special geometrical properties of network data, we design a sign-constrained least square procedure to find the aggregation, which significantly improves the strategy. Our method is much simpler to implement and possesses strong theoretical guarantees.  Our numerical and theoretical analyses show that the proposed mixing methods nearly achieve the oracle estimators' accuracy regardless of whether the true model is contained in the candidate model set. Furthermore, an evaluation of 385 real-world networks  shows that the proposed methods are more accurate and much more computationally efficient than the stacking method of \cite{ghasemian2020stacking}, which still lacks theoretical support.

The rest of the paper is organized as follows. Section~\ref{sec:method} introduces the proposed methods and their theoretical properties. We start from the basic setup of the mixing strategy. Exponential mixing is first introduced as a warm-up, which is shown to achieve single-model adaptivity. Then we introduce the more general non-negative linear mixing method and show that the non-negative property is crucial in network settings.  Section~\ref{sec:simulation} includes extensive simulation studies of the (almost) tuning-free properties and compares the mixing estimators with several benchmark methods under both graphon (nonparametric) models and parametric models. In Section~\ref{sec:data}, we consider link prediction tasks on 385 real-world networks from 6 domains, introduced in \cite{ghasemian2020stacking}. The evaluation demonstrates the advantage of the proposed method as an ``off-the-shelf" method for link prediction. Section~\ref{sec:disc} concludes the paper.

%% file: Setup.tex
\paragraph{Notations.} Given a matrix $M$, we use $\norm{M}$ and $\norm{M}_F$ to denote its spectral norm and Frobenius norm, respectively, and let $\norm{M}_{\infty} = \max_{ij}|M_{ij}|$. In addition, for any two matrices $P_1, P_2$ of the same dimensions, we define $\langle P_1, P_2 \rangle = \text{trace}(P_1^TP_2)$. Given a matrix $M\in \bR^{n\times n}$ and an index set $\Omega \subset [n]\times [n]$, denote by $M(\Omega)$  the matrix obtained from $M$ by setting all entries of $M$ with indices in the complement $\Omega^c$ of $\Omega$ to zero. For two sequences $\{a_n\}$ and $\{b_n\}$, we write $a_n\lesssim b_n$ if $a_n = O(b_n)$, in which case we can also write it as $b_n \gtrsim a_n$. If $a_n \lesssim b_n$ and $a_n \gtrsim b_n$, we write  $a_n \approx b_n$. We say that an event $E$ occurs with high probability if $\mathbb{P}(E)\ge 1-n^{-\delta}$ for some constant $\delta>0$. We use $C>0$ to denote an absolute constant whose value may change from line to line.

%In this paper, we assume the network to be undirected, unweighted without self-loops, even though the generalization of our method to other settings would not be difficult.
%We consider the inhomogeneous Erd\H{o}s-R\'{e}nyi network model as our probabilistic framework:  There exists a probability matrix $P \in \bR^{n\times n}$ such that the above-diagonal entries $A_{ij}$ of $A$ are independent Bernoulli random variables with probabilities $P_{ij}$. 
We focus on unweighted undirected networks in this paper. Let $n$ be the network size. A network can be represented by a binary adjacency matrix $A \in \{0,1\}^{n\times n}$, where $A_{ij} = 1$ if and only if node $i$ and node $j$ are connected. Since the network is undirected, we have $A = A^T$. Furthermore, assume $A_{ii}=0$ for all $i$ (we do not consider self-loops). Our discussion of networks will be embedded in the  \emph{inhomogeneous Erd\H{o}s-R\'{e}nyi network model} \citep{Bollobas2007}, which requires that the upper diagonal entries of $A$ are independent Bernoulli random variables. Denote $P=\e A$ and assume that there exists a set $\mathcal{M}$ of $m$ available methods for estimating $P$ from $A$. These may be parametric estimation procedures, such as those based on the stochastic block model \citep{chen2014network, li2016network} or the latent space model \citep{ma2020universal}, or nonparametric algorithms \citep{chatterjee2015matrix, airoldi2013stochastic, chan2014consistent, gao2015rate,zhang2015estimating}.  Our goal is to leverage these methods to derive a new estimator $\hat{P}$ that can achieve a better approximation of $P$ in terms of the mean squared error.    A natural baseline target is, not surprisingly, to ensure that $\hat{P}$ is as good as the unknown optimal estimator in $\mathcal{M}$. Such a property is often referred to as \emph{adaptivity} \citep{catoni1997mixture,yang2000mixing,yang2001adaptive}. Ultimately, our goal will be more ambitious than adaptivity: we would like to achieve optimality within a broader class of estimators than the $m$ estimators of $\mathcal{M}$. We aim for an off-the-shelf option for modeling networks. That is, in addition to strong theoretical properties, empirical applicability is equally important. In a linear regression setting, procedures with provable adaptivity may not be computationally feasible or may require tuning parameters \citep{catoni1997mixture,yang2000mixing,yang2001adaptive,Tsybakov2003,bunea2007aggregation,vanderLaan.superlearner.2007}, but we will introduce a computationally efficient and tuning-free method that can be scalable even to large networks.

Motivated by the adaptive mixing method in the classical regression setting, we resort to a data-splitting strategy to design our method. The strategy is based on the dyad-splitting of the network cross-validation procedure of \cite{li2016network}. Specifically, given a fixed $p\in(0,1)$, we randomly choose a subset of node pairs $\Omega\subset [n]\times [n]$ by independently selecting each node pair with probability $p$; in our setting, $p$ is usually a constant between 0 and $0.5$. Then $A(\Omega^c)$ can be viewed as an adjacency matrix generated from the inhomogeneous Erd\H{o}s-Renyi model with probability matrix $(1-p)P$.  Therefore, any model fitting methods for inhomogeneous Erd\H{o}s-Renyi networks can be used with the same theoretical guarantees for consistency as their original version as discussed in \cite{gao2020discussion} and \cite{li2020rejoinder}. We apply the methods from $\mathcal{M}$ to $A(\Omega^c)$ to get estimators of $(1-p)P$, and multiply them by $(1-p)^{-1}$ to obtain estimators $\hat{P}^{(1)},...,\hat{P}^{(m)}$ of $P$. We assume that the entries of these estimators always fall between 0 and 1. If this is not the case, we can always threshold the entries without increasing the error. The mixing strategy is essentially an aggregated version of the $m$ estimators for $P$.  

We use the hold-out entries $A(\Omega)$  to determine a reasonable way to combine the available estimators. A potential approach for this purpose is the cross-validation method of \cite{li2016network}, which calculates $
\|A(\Omega)-\hat{P}^{(r)}(\Omega)\|_F^2,  1\le r\le m
$, and picks the one with the smallest error:
\begin{equation}\label{eq:ECV-estimator}
\hat{P} = \hat{P}^{(\hat{r})}, \quad \hat{r} = \argmin_{1\le r\le m}\|A(\Omega)-\hat{P}^{(r)}(\Omega)\|_F^2.
\end{equation}
(In practice, one often repeats this procedure multiple times and uses the average validation error as the criterion. Here we focus on the single split for conceptual simplicity.) 
This is one of the most common model selection strategies in statistical modeling \citep{hastie2009elements}.  
%By using the model selection, we hope to identify the best model for fitting the data. 
It has been verified, for example, in the context of multivariate outcome prediction and density estimation \citep{vanderLaan.crossvalidation.2006,feng2019restricted,lei2020cross}. But so far it is unclear whether such an approach is valid for network cross-validation. For example, although the estimator in \eqref{eq:ECV-estimator} works well for moderately dense networks when $\mathcal{M}$ contains the true model, its performance deteriorates quickly in a sparse regime or under model misspecification (as we show in Section~\ref{sec:simulation}). We now introduce a soft-selection version of the cross-validation method to address this shortcoming.  
%The proposed estimator is a linear combination of the available estimators with exponential weights and theoretically matches the best individual model adaptivity. 

\subsection{Network mixing with exponential weights and estimation adaptivity}\label{sec:model selection}

We view the validation error $\|A(\Omega)-\hat{P}^{(r)}(\Omega)\|_F^2$ as a goodness-of-fit metric for the $r$th model. To match the performance of the optimal model in $\mathcal{M}$, we focus on the models with small validation errors. In particular, we propose the following simple rule to combine the models based on  exponential weights:
\begin{eqnarray}\label{eq:exp weights}
\hat{P}^{\text{(exp)}} = \sum_{r=1}^m \pi_r \hat{P}^{(r)}, \qquad \pi_r= \frac{ \exp\left(-\|A(\Omega)-\hat{P}^{(r)}(\Omega)\|_F^2\right)}{ \sum_{r=1}^m\exp\left(-\|A(\Omega)-\hat{P}^{(r)}(\Omega)\|_F^2\right)}.
\end{eqnarray}
Compared with \eqref{eq:ECV-estimator}, this estimator is a soft-selection version of the cross-validation procedure. Despite its simplicity,  we now show that it achieves  model adaptivity, matching the performance of the unknown optimal estimator produced by $\mathcal{M}$. Since $\Omega$ is independent of the data, all theoretical analysis is conditioned on $\Omega$. The data-splitting proportion $p$ is a fixed constant, independent of the network size $n$.

\begin{thm}[Mixing estimator with exponential weights]\label{thm:exp weights}
Let $A$ be the adjacency matrix of a random network drawn from the inhomogeneous Erd\H{o}s-R\'{e}nyi model with probability matrix $P=\mathbb{E} A$. 
%For a fixed $p\in(0,1)$, let $\Omega$ be the index set obtained by independently selecting elements of $[n]^2$ with probability $p$, and $\hat{A}$ be the matrix obtained from $A$ by setting entries of $A$ with indices in $\Omega$ to zero. 
Denote by $\hat{P}^{(r)}$, $1\le r \le m$, the estimators of $P$ obtained by applying the methods from $\mathcal{M}$ to $A(\Omega)$, and let $\hat{P}$ be the convex combination of these estimates defined by \eqref{eq:exp weights}. Assume $m = o(n^2)$. Then with high probability,
\begin{equation}\label{eq:exp-error}
\big\|\hat{P}^{\text{(exp)}}(\Omega)-P(\Omega)\big\|_F \le \min_{1\le r\le m}\big\|\hat{P}^{(r)}(\Omega)-P(\Omega)\big\|_F+ C\varepsilon,
\end{equation}   
where $C>0$ is a constant and   
\begin{eqnarray*}
\varepsilon = \Big(\log n\cdot\max_{i,j}P_{ij}\Big)^{1/2}+\min\left\{\frac{\log(nm)}{\min_{1\le r\le m}\|P^{(r)}(\Omega)-P(\Omega)\|_F}, \ \log^{1/2}(nm)\right\}.
\end{eqnarray*}
\end{thm}

\begin{rem}
Although Theorem~\ref{thm:exp weights} describes the error restricted to $\Omega$, extending the error bound to the entire matrix $P$ is trivial. Replacing $\Omega$ by $\Omega^c = [n]^2\setminus\Omega$ in the estimating procedure and applying Theorem~\ref{thm:exp weights} would generate another estimator $\breve{P}$ that admits the same type of error bound. Then combining $\hat{P}(\Omega)$ and $\breve{P}(\Omega^c)$ as a full estimator would produce the same error bound for the entire matrix $P$. For simplicity, we will state our results only in terms of $P(\Omega)$ in all theoretical discussions to follow.
\end{rem}

Theorem~\ref{thm:exp weights} states that the estimator given in \eqref{eq:exp weights} is nearly as good as the best estimate produced by a single method from $\mathcal{M}$. To better understand the additional error $\varepsilon$, assume that the network is generated from a stochastic block model, under which nodes are partitioned into $k$ groups according to a label vector $c\in[n]^k$ and edges are formed independently between nodes with probabilities $P_{ij}=B_{c_ic_j}$ for some fixed matrix $B\in\mathbb{R}^{k\times k}$. According to \cite{gao2015rate},
\begin{eqnarray*}
\min_{1\le r\le m}\big\|\hat{P}^{(r)}(\Omega)-P(\Omega)\big\|_F^2 \gtrsim k^2 + n\log k, 
\end{eqnarray*}     
while the square of the additional error is
\begin{eqnarray*}
\varepsilon^2 \lesssim \log n + \min\left\{\frac{\log^2(nm)}{k^2+n\log k},\log(nm)\right\} \approx \log n + \frac{\log^2(nm)}{k^2+n\log k}. 
\end{eqnarray*}
It is easy to see that $\varepsilon^2$ is smaller than the rate-optimal error $k^2 + n\log k$. Therefore, our estimator can always match the optimal estimator from $\mathcal{M}$.

As a connection to similar properties of statistical estimation problems, \cite{catoni1997mixture} and \cite{yang2000mixing,yang2001adaptive} introduce mixing methods for regression and density estimation that can achieve estimation adaptivity. Unfortunately, their strategies are computationally expensive even in regression settings, let alone in network problems, the sample size of which scales in $O(n^2)$.   In contrast, our estimator \eqref{eq:exp weights}  achieves the same type of adaptivity with negligible computational cost in addition to the $m$ model-fitting procedures.

\subsection{Beyond adaptivity: Non-negative linear network mixing} 

The exponential mixing estimator \eqref{eq:exp weights}  performs well when at least one of the $m$ individual estimators is accurate. In practice, however, all the available estimators may be inaccurate, for example, when the network is very sparse. Given the $m$ estimators at hand, it is natural to be more ambitious: to seek a better candidate than any single model estimate.

Consider the class of linear combinations of the $m$ estimators
\begin{equation}\label{eq:generic-estimator}
\hat{P} = \sum_{r=1}^m \pi_r \hat{P}^{(r)}.
\end{equation}
The most natural option is the \emph{linear mixing estimator}, which uses the weights provided by solving the ordinary least squares (OLS) problem:
\begin{eqnarray}\label{eq:ols optimization}
\hat{\pi}^{\text{(ols)}} = \argmin_{\pi \in \bR^m}\Big\|\sum_{r=1}^m \pi_r \hat{P}^{(r)}(\Omega) - A(\Omega)\Big\|_F^2. 
\end{eqnarray}    

Although linear mixing performs very well in many settings, it is a generic method and does not leverage many features of  network estimation. For example, all the entries of $P$ are non-negative, and therefore it is expected that $\langle \hat{P}^{(r)}(\Omega),P(\Omega)\rangle>0$ for any reasonable estimate $\hat{P}^{(r)}(\Omega)$. Similarly, $\langle \hat{P}^{(r)}(\Omega),\hat{P}^{(s)}(\Omega) \rangle > 0$ for all $1\le r,s\le m$ (unless their supports are disjoint), and the angles between reasonably good estimators tend to be small. Moreover, in sparse networks the noises (entries of $A-P$) are heavy-tailed random variables, so the available estimators of $P$ can be very noisy. These observations motivate us to further improve the linear mixing method by introducing a natural non-negative constraint to the OLS weights.    
%\begin{itemize}
%\item For any reasonable methods, $\langle \hat{P}^{(r)}(\Omega),P(\Omega) \rangle  >  0$, $1\le r \le m$.
%\item For any set of estimators, we have
%$$\langle \hat{P}^{(r)}(\Omega),\hat{P}^{(s)}(\Omega) \rangle > 0, 1\le r,s \le m.$$
%And the angle tend to be estimators (after vectorization) tend to be small for many pairs of estimators.
%\item The concentration of $A(\Omega)$ around $P(\Omega)$ tend to be worse when the network is sparse. So the estimation tend to be noisy.
%\end{itemize}
%The above observations indicate that it may be reasonable to introduce constraints on the model space for mixing, and the non-negativity of mixing weights can be potentially a good option. 
In particular, we consider the \emph{non-negative linear} (NNL) mixing estimator based on the following weights:
\begin{eqnarray}\label{eq:nnls optimization}
\hat{\pi}^{\text{(nnl)}} = \argmin_{\pi \succeq 0}\Big\|\sum_{r=1}^m \pi_r \hat{P}^{(r)}(\Omega) - A(\Omega)\Big\|_F^2, 
\end{eqnarray}    
where $\pi \succeq 0$ denotes the constraint that all the entries of the weight vector $\pi$ are non-negative. This is a simple convex optimization problem and can be solved efficiently by either a projected quasi-Newton algorithm or a sequential coordinate descent algorithm \citep{kim2006new, chen2010nonnegativity}.
The NNL mixing estimator is defined to be
\begin{eqnarray}\label{eq:nnls estimate}
\hat{P}^{\text{(nnl)}} = \sum_{r=1}^m \hat{\pi}_r^{\text{(nnl)}} \hat{P}^{(r)}.
\end{eqnarray}    
We can see that $\hat{P}^{\text{(nnl)}}(\Omega)$ is the projection of $A(\Omega)$ on the convex cone formed by the conical combination of $m$ individual estimators $\hat{P}^{(r)}(\Omega), 1\le r \le m$. The non-negative sign constraint directly imposes a regularization effect that significantly helps the method handle the potentially large number of individual estimators. This is crucial for a tuning-free procedure and matches our aim for an ``off-the-shelf" method.

The non-negative constraint has proved effective in high-dimensional linear regression problems \citep{Slawski.NNLS.2011, slawski2013non, meinshausen2013sign}, with properties similar to the LASSO estimator. However, in the current context, NNL estimation is not motivated by the curse of dimensionality. In network mixing problems, the sample size for \eqref{eq:ols optimization} is of order $n^2$, and $m$ is usually much smaller than $n^2$.  Therefore, though $m$ can be large if one wants to ensure the expressiveness of the mixing estimator, \eqref{eq:ols optimization} is usually not an ultra-high-dimensional problem. Instead, the strong correlation between the $ \hat{P}^{(r)}$s and the low concentration of the adjacency matrix (due to the network sparsity) complicate the matter and result in the deterioration of the OLS estimator, even in a relatively low-dimensional setting. 

\begin{figure}[H]
\centering
\begin{subfigure}[t]{0.3\textwidth}
\centering
\includegraphics[width=\textwidth]{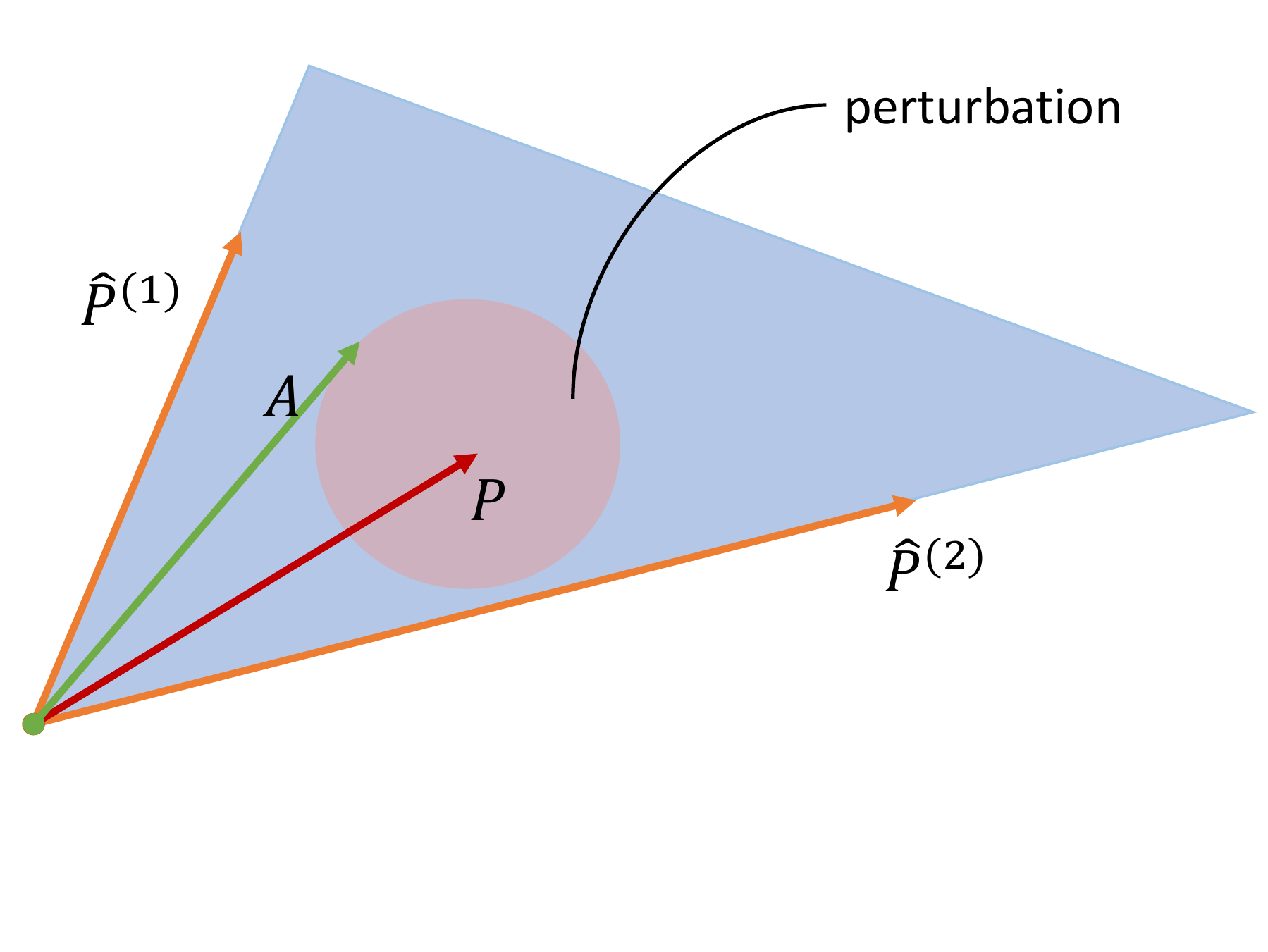}
\caption{Good concentration and weak alignment}
\label{fig:demo-normal}
\end{subfigure}
\hfill
\begin{subfigure}[t]{0.3\textwidth}
\centering
\includegraphics[width=\textwidth]{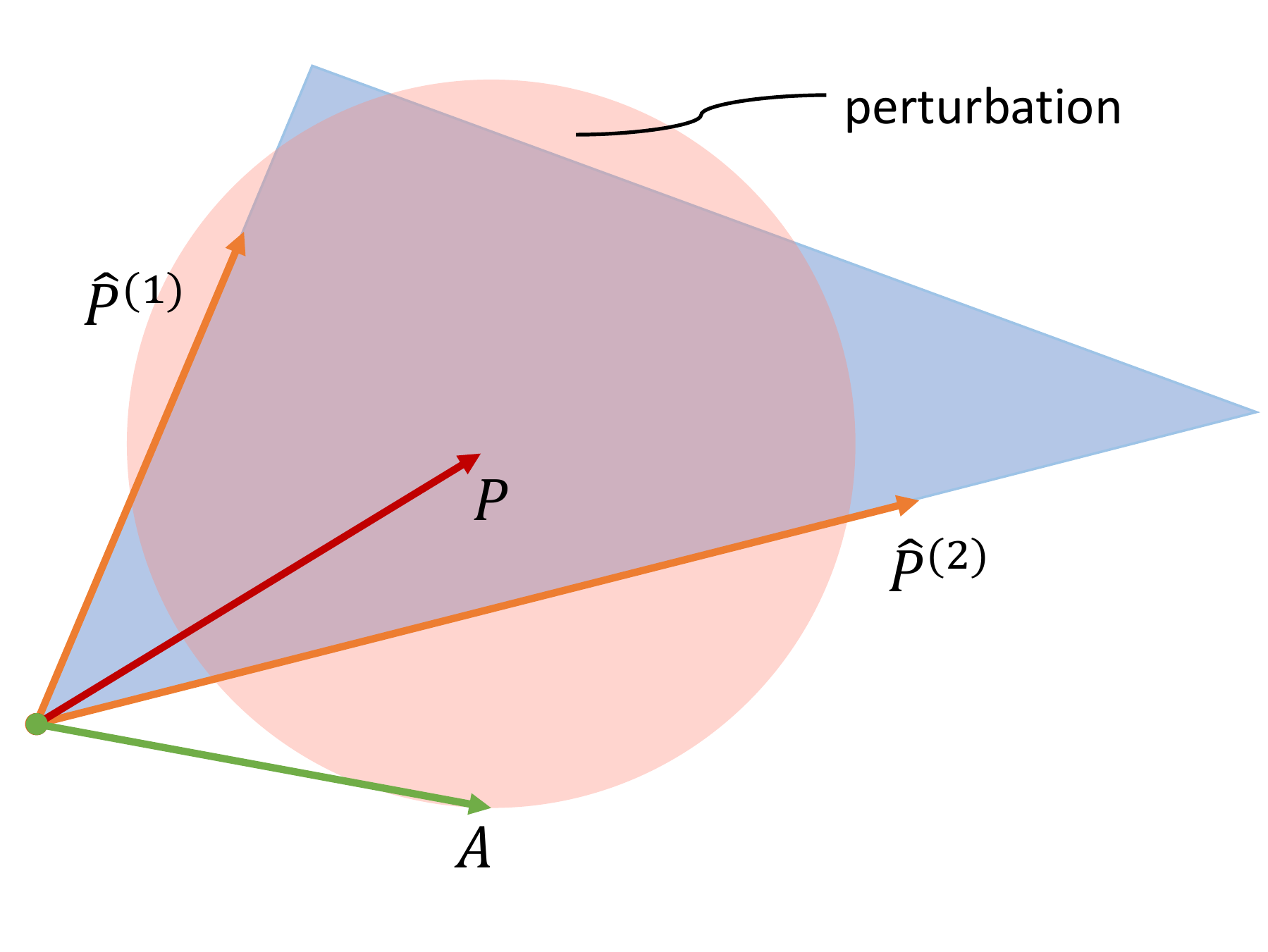}
\caption{ Bad concentration and weak alignment}
\label{fig:demo-correlation}
\end{subfigure}
\hfill
\begin{subfigure}[t]{0.3\textwidth}
\centering
\includegraphics[width=\textwidth]{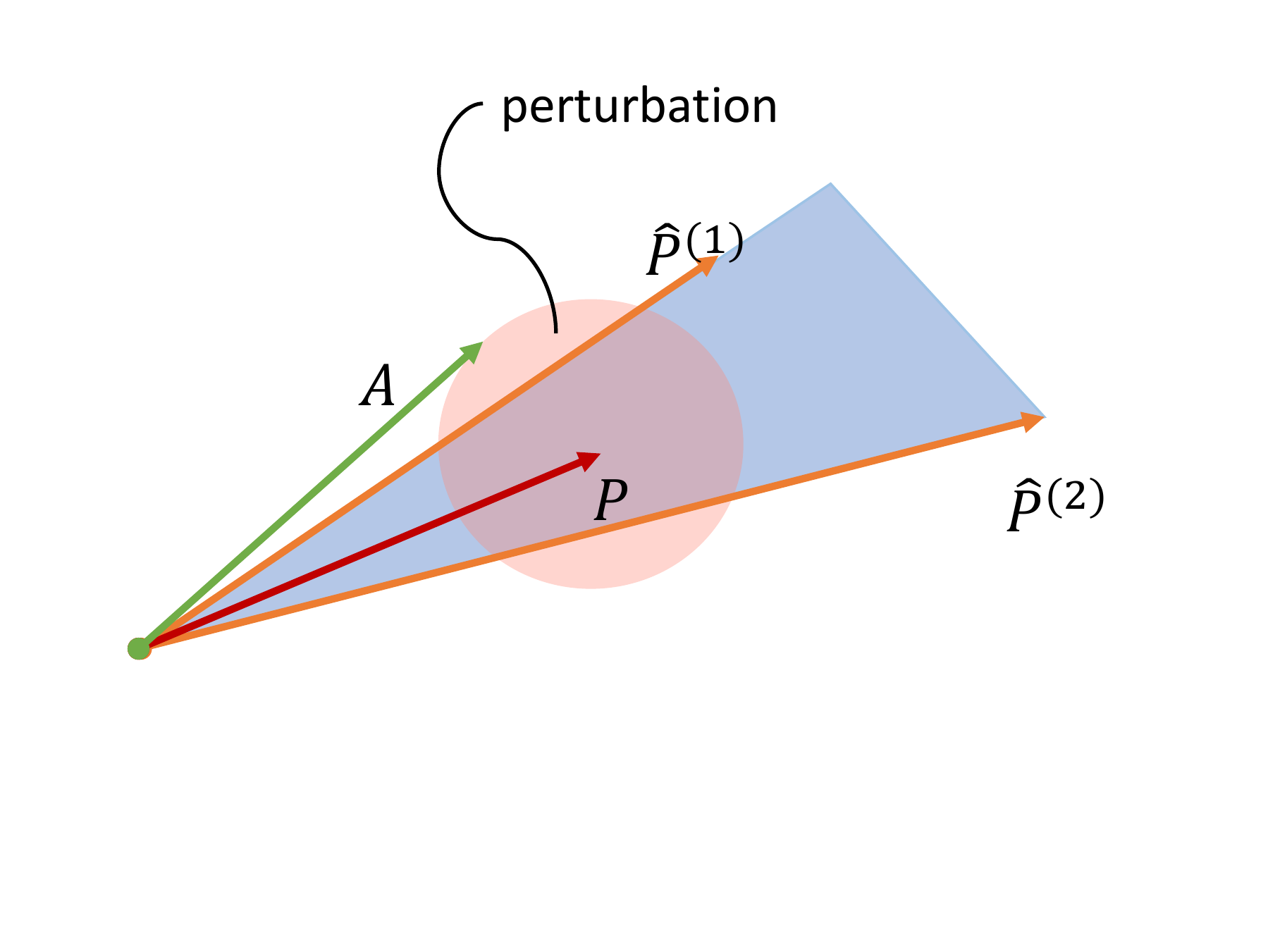}
\caption{Good concentration and strong alignment}
\label{fig:demo-sparse}
\end{subfigure}
\caption{Illustration of the relation between the convex cone and the concentration of the adjacency matrix.}
\label{fig:demo}
\end{figure}

To see why the non-negative constraint can help, let us assume that  the true signal $P(\Omega)$ is within the convex cone of the estimators. Figure~\ref{fig:demo-normal} illustrates an ideal situation. When the two individual estimators form a large angle, and the perturbation range of $A$ around $P$ is not too large, the observed adjacency matrix is likely to stay within or close to the convex cone, so the OLS and NNL estimators perform similarly. When the perturbation range is large, as in Figure~\ref{fig:demo-correlation} (which happens in sparse networks \citep{le2017concentration}), $A$ may be far from the cone (and $P$), and the projection to the cone can result in a better estimator.
On the other hand, when the two estimators align well, the convex cone is smaller (Figure~\ref{fig:demo-sparse}), and $A$ is unlikely to belong to the cone. In this case, the NNL estimator can also outperform the OLS estimator. Overall, these observations show that sparse networks and strong alignment between estimators tend to make the OLS estimator  vulnerable to  increasing $m$, even when $m$ is much smaller than $n^2$.

%It is interesting to compare the NNLS estimate and the ordinary least square (OLS) estimate for which the weights are the solution of the least square problem without the non-negative constraint. Our numerical analysis shows that the NNLS estimate usually outperforms the OLS estimate for many simulated and real networks. This is partly because when the set of available methods $\mathcal{M}$ is large, which is a common scenario for model aggregation, the solution of the NNLS problem is often much sparser than the OLS solution. Consequently, the NNLS estimate does not suffer from the spurious errors incurred by the OLS estimate. It is also often the case that estimates produced by the methods in $\mathcal{M}$ are highly correlated, rendering the OLS estimate unstable. In contrast, the NNLS remains robust and performs very well in this setting. Our theoretical results in the next section further support these observations.            

We now present the theoretical support for the NNL estimator and its claimed advantage over linear mixing. Let
\begin{eqnarray*}
\mathcal{L}=\Big\{\sum_{r=1}^m \pi_r\hat{P}^{(r)}(\Omega), \pi_r\in\mathbb{R},1\le r \le m\Big\}, \quad \mathcal{C}=\Big\{\sum_{r=1}^m\pi_r\hat{P}^{(r)}(\Omega), \pi_r\ge 0, 1\le r \le m\Big\}
\end{eqnarray*}
be the linear subspace and the cone generated by $\{\hat{P}^{(r)}(\Omega),1\le r\le m\}$. Denote by $\Pi_\mathcal{L}$ and $\Pi_\mathcal{C}$ the projections onto $\mathcal{L}$ and $\mathcal{C}$, respectively. 

\begin{thm}[Non-negative mixing estimator, positive inner products]\label{thm:nnls}
Let $A$ be the adjacency matrix of a random network drawn from the inhomogeneous Erd\H{o}s-R\'{e}nyi model with probability matrix $P=\mathbb{E} A$. Denote by $\hat{P}^{(r)}$, $1\le r \le m$, the estimates of $P$ obtained by applying the methods from $\mathcal{M}$ to $A(\Omega^c)$, and let $\hat{P}^{\text{(nnl)}}$ be the NNL estimator defined by \eqref{eq:nnls estimate}. Assume that
\begin{equation}\label{eq:correlated assumption}
\delta  = \min_{1\le r,s\le m}\frac{\langle \hat{P}^{(r)}(\Omega),\hat{P}^{(s)}(\Omega) \rangle}{\|\hat{P}^{(r)}(\Omega)\|_F\cdot\|\hat{P}^{(s)}(\Omega)\|_F}>0.
\end{equation}
Then there exists a constant $C>0$ such that with high probability,
\begin{eqnarray}\label{eq:nnls bound}
\|\hat{P}^{\text{(nnl)}}(\Omega)-P(\Omega)\|_F &\le& \|\Pi_\mathcal{C}P(\Omega)- P(\Omega)\|_F + C\delta^{-1/2}\varepsilon, 
\end{eqnarray} 
where $C>0$ is a constant and 
\begin{equation}\label{eq:nnls-tail}
\varepsilon = \max_{1\le r \le m}\frac{\|\hat{P}^{(r)}(\Omega)\|_\infty}{\|\hat{P}^{(r)}(\Omega)\|_F}\cdot\log(n+m) +\sqrt{\|P(\Omega)\|_\infty\log(n+m)}.
\end{equation}
\end{thm}

Theorem~\ref{thm:nnls} states that the proposed estimator $\hat{P}^{\text{(nnl)}}(\Omega)$ is nearly as accurate as the non-negative linear oracle $\Pi_\mathcal{C}P(\Omega)$. Given a fixed $\delta$, the extra error term $\varepsilon$ can be at most of the order $\log(n+m)$, although it will grow much more slowly if the entries of $P$ and $\hat{P}^{(r)}$ are of similar orders and the network is relatively sparse.        

As a consequence of Theorem~\ref{thm:nnls}, the next corollary shows that the NNL estimator can be strictly more accurate than the OLS estimator $\Pi_\mathcal{L} A(\Omega)$, especially when $m$ is large (a typical scenario in model aggregation).  

\begin{coro}[Comparison of NNL and OLS estimators]\label{cor:comparison}
Let $A$ be the adjacency matrix of a random network drawn from the inhomogeneous Erd\H{o}s-R\'{e}nyi model with probability matrix $P=\mathbb{E} A$. Denote by $\hat{P}^{(r)}$, $1\le r \le m$, the estimators of $P$ obtained by applying the methods from $\mathcal{M}$ to $A(\Omega^c)$, and let $\hat{P}^{\text{(ols)}}$ be the OLS estimator defined by \eqref{eq:ols optimization}. With high probability, 
\begin{equation}\label{eq:ols-oracle}
\norm{\hat{P}^{(\text{ols})}(\Omega) - P(\Omega)}_F^2 \le \norm{\Pi_{\mathcal{L}}P(\Omega) - P(\Omega)}_F^2 + m\|P(\Omega)\|_\infty + \left(m\log n\|P(\Omega)\|_\infty\right)^{1/2}. 
\end{equation}
Furthermore, consider the setting in Theorem~\ref{thm:nnls} and denote 
$$\rho = \big(1-\|P(\Omega)\|_\infty\big)^2\cdot\min_{(i,j)\in\Omega} P_{ij}^2, \quad \Delta = \|\hat{P}^{\text{(ols)}}(\Omega)-P(\Omega)\|_F^2 - \|\hat{P}^{\text{(nnl)}}(\Omega)-P(\Omega)\|_F^2.$$ 
Assume $m \ge C\|P(\Omega)\|_\infty\rho^{-1}\log n$ for some sufficiently large constant $C>0$. Then with high probability,
\begin{equation}\label{eq:Delta bound}
\Delta \ge \frac{m\rho^{1/2}}{2} - \Psi, 
\end{equation} 
where 
\begin{eqnarray*}
\Psi = \frac{C'\log^2(n+m)}{\delta}\Big(\max_{1\le r \le m}\frac{\|\hat{P}^{(r)}(\Omega)\|_\infty^2}{\|\hat{P}^{(r)}(\Omega)\|_F^2} +\|P(\Omega)\|_\infty\Big)+ \|\Pi_\mathcal{C}\Pi_\mathcal{L} P(\Omega)-\Pi_\mathcal{L} P(\Omega)\|_F^2
\end{eqnarray*}
for some constant $C'>0$.
\end{coro}

Note that the term $\|\Pi_\mathcal{C}\Pi_\mathcal{L} P(\Omega)-\Pi_\mathcal{L} P(\Omega)\|_F$ is the distance from the linear oracle $\Pi_\mathcal{L}P(\Omega)$ to the cone $\mathcal{C}$, measuring the degree of violation of the non-negative cone assumption. Corollary~\ref{cor:comparison} demonstrates the effects of the main factors on performance. It can be seen that the advantage of NNL mixing over OLS mixing increases with $m$ and also with the correlation $\delta$, as we have discussed above. Next, we illustrate the effect of network density.

\begin{ex}\label{rem:degree}
Assume a sparse network and that all the entries of $P$ are of the same order $d/n = o(1)$, where $d$ is the average degree. Also, assume that the $m$ individual estimators are reasonable, so all the $\hat{P}^{(r)}_{ij}$s are also of the same order as the $P_{ij}$s. When the conical assumption holds and $m \gg n\log{n}/d$, taking the relative error to cancel the scaling effect of $P$, we have
$$\frac{\Delta}{\norm{P(\Omega)}_F^2}  \gtrsim \frac{m}{nd} - \frac{\log^2(n+m)}{\delta d^2}.$$
For sparser networks with small $d$, the advantage of the NNL estimator is more dramatic. This theoretical prediction is further supported by our empirical evidence in Section~\ref{sec:simulation}.
\end{ex}

Theorem~\ref{thm:nnls} requires $\hat{P}^{(r)}(\Omega)$, $1\le r\le m$, to have positive inner products. This assumption is reasonable because they are all estimators of $P(\Omega)$ and must have non-negative entries. More importantly, we can directly calculate $\delta$ from data. 

Next, we consider an even weaker assumption, which is true in all reasonable settings we can think of. We replace the assumption \eqref{eq:correlated assumption} in Theorem~\ref{thm:nnls} with the following assumption on $\hat{P}^{(r)}(\Omega)$, $1\le r\le m$, known as the {\em self-regularizing property} \citep{Slawski.NNLS.2011}. Denote by $\Sigma\in\mathbb{R}^{m\times m}$ the matrix with entries $\Sigma_{rs}=\langle \hat{P}^{(r)}(\Omega), \hat{P}^{(s)}(\Omega)\rangle$ and $\|\Sigma\|_\infty=\max_{1\le r,s\le m}|\Sigma_{rs}|$. We say that $\Sigma$ satisfies the self-regularizing property with constant $\kappa$,  $0<\kappa\le 1$, if 
\begin{eqnarray}\label{eq:self-reg}
\beta^T\Sigma\beta \ge \kappa\|\Sigma\|_\infty, \text{ for all $\beta\succeq 0$  and $\sum_{r=1}^m\beta_r = 1$}.
\end{eqnarray}
Notice that, like \eqref{eq:correlated assumption}, this condition can be numerically verified from the data by solving a convex problem. More importantly, in our current setting of network mixing, this condition will almost always hold due to the following property, taken directly from the discussion of \cite{Slawski.NNLS.2011}.

\begin{prop}\label{prop:self-reg}
If there exists a partition $\{ Q_t\}_{t=1}^T$ of $[m]$ such that 
$$\min_{r, s \in Q_t} \langle \hat{P}^{(r)}(\Omega), \hat{P}^{(s)}(\Omega)\rangle \ge \kappa \max_{r\in Q_t}\norm{\hat{P}^{(r)}(\Omega)}_F^2 >0, \text{ for all } 1\le t \le T,$$
then $\Sigma$ is self-regularizing with constant $\kappa/T$.
\end{prop}

To see why \eqref{eq:self-reg} is weaker than  \eqref{eq:correlated assumption}, notice that without loss of generality we can assume that all the $\hat{P}^{r}(\Omega)$s have the same norm, because rescaling does not change our linear fitting. So when \eqref{eq:correlated assumption} holds, $\Sigma$ also satisfies \eqref{eq:self-reg}, with $\kappa = \rho$. Moreover, even if we are in the extreme situation where
$$\langle \hat{P}^{(r)}(\Omega), \hat{P}^{(s)}(\Omega)\rangle = 0$$
for some $r, s$, Proposition~\ref{prop:self-reg} indicates that \eqref{eq:self-reg} can still hold as long as we separate such pairs in different groups. Even in the worst case, the $m$-way partition guarantees the self-regularizing property, with $\kappa = 1/m$. 

\begin{thm}\label{thm:self-regularizing}
Consider the setting of Theorem~\ref{thm:nnls} with condition \eqref{eq:correlated assumption} replaced by the self-regularizing property \eqref{eq:self-reg}. Then there exists a constant $C>0$ such that with high probability,
\begin{eqnarray*}
\|\hat{P}^{\text{(nnl)}}(\Omega)-P(\Omega)\|_F &\le& \inf_{\eta \in \ccal}\|\eta-\Pi_\mathcal{L} P(\Omega)\|_F+ C\kappa^{-1/2}\epsilon',
\end{eqnarray*}
where 
\begin{eqnarray*}
\epsilon' &=& \left(\frac{\Phi\log(n+m)}{\sqrt{\|\Sigma\|_\infty}} +\sqrt{\Phi\|\pi^*\|_1\log(n+m)}\right),\\
\Phi &=& \max_{1\le r \le m}\|\hat{P}^{(r)}(\Omega)\|_\infty + \|P(\Omega)\|_\infty^{1/2}\norm{\Sigma}_{\infty},
\end{eqnarray*}
and $\pi^*$ is the non-negative linear coefficient of the oracle estimator such that 
$$\Pi_{\ccal}P(\Omega) = \sum_{r}\pi^*_r\hat{P}^{(r)}(\Omega).$$
\end{thm}

Compared with Theorem~\ref{thm:nnls}, the price we pay for the weaker assumption in Theorem~\ref{thm:self-regularizing} is greater additive error. As a simple demonstration, consider the setting of Example~\ref{rem:degree} and assume that all the $\hat{P}^{r}(\Omega)$s have the same norm. In this case, $\kappa = \delta$. At best, the $\epsilon'$ of Theorem~\ref{thm:self-regularizing} has
$$\epsilon' \approx \frac{d^{3/2}}{n^{1/2}}\log(n+m) + \frac{d^{5/4}}{n^{1/4}}\log^{1/2}(n+m). $$
In contrast, the $\epsilon$ of Theorem~\ref{thm:nnls} gives
$$\epsilon \approx \frac{1}{n}\log(n+m) + \frac{d^{1/2}}{n^{1/2}}\log^{1/2}(n+m), $$
which is clearly of a lower order.

\subsection{Candidate set and other practical considerations}\label{secsec:practical}

The mixing strategy involves determining a set of candidate estimators and the hold-out proportion, $p$. For the OLS and NNL mixing methods, the scales of the individual estimators do not matter, and they do not need to be accurate individually. This property is essential for a mixing method to be applicable for general link prediction problems (see Section~\ref{sec:data}). But exponential mixing does not make sense if the $\hat{P}^{(r)}$s are in the wrong scale or if they are all inaccurate. 

In determining the candidate set $\mcal$, there is a necessary tradeoff between computational efficiency and expressiveness. We recommend using spectral clustering together with SBM fitting for $1 \le k \le K_{\max}$, spherical spectral clustering \citep{rohe2011spectral,lei2014consistency}  and DCBM fitting for $1 \le k \le K_{\max}$, and the universal singular value thresholding (USVT) of \cite{chatterjee2015matrix}, with the improvement mentioned in \cite{zhang2015estimating}, by taking the first $n^{1/3}$ singular components. This set of candidate estimators can be calculated efficiently for large networks, with the main computational burden being the one-time calculation of SVD. The block models, despite their simplicity, have been shown to have great approximation power if they are used properly \citep{airoldi2013stochastic}, and USVT also comes with good expressiveness. When networks are sparse, the block components likely receive higher mixing weights, helping stabilize the estimator.
In contrast, when networks are sufficiently dense, USVT can be more effective. Empirically, we observe that this setup gives very effective estimation performance. In the above recommendation, $K_{\max}$ is a reasonably large positive integer, and in Sections~\ref{sec:simulation} and \ref{sec:data}, we show that our method is very robust to $K_{\max}$ and $\rho$ within reasonable ranges, which is again due to the regularizing property of the non-negative constraint.

The discussion so far has focused on one random split, $\Omega$. In practice, one can also randomly repeat the procedure multiple times and take the average as the output. This may slightly improve accuracy in our limited evaluation but will also increase the computational cost by a multiplicative factor. For this reason, we only consider single splits in this study.

The mixing method, and any aggregation methods, do have limitations. The primary performance metric of the mixing approach is estimation accuracy. Combining multiple estimators may destroy the structural interpretations of each individual estimator, such as block structures or smoothness. However, when special structures are desirable, one can always apply the same structural extraction strategy, such as community detection, to the estimated $\hat{P}$. A more accurate estimate of $P$ may lead to more accurate structure extraction.

%% file: Simulation.tex
\begin{figure}[H]
\centering
\begin{subfigure}[t]{0.3\textwidth}
\centering
\includegraphics[width=\textwidth]{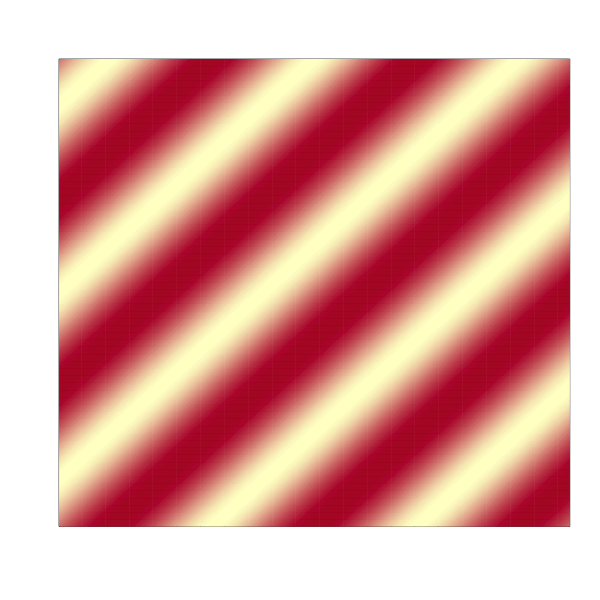}
\caption{Model 1}
\label{fig:bars-graphon}
\end{subfigure}
\hfill
\begin{subfigure}[t]{0.3\textwidth}
\centering
\includegraphics[width=\textwidth]{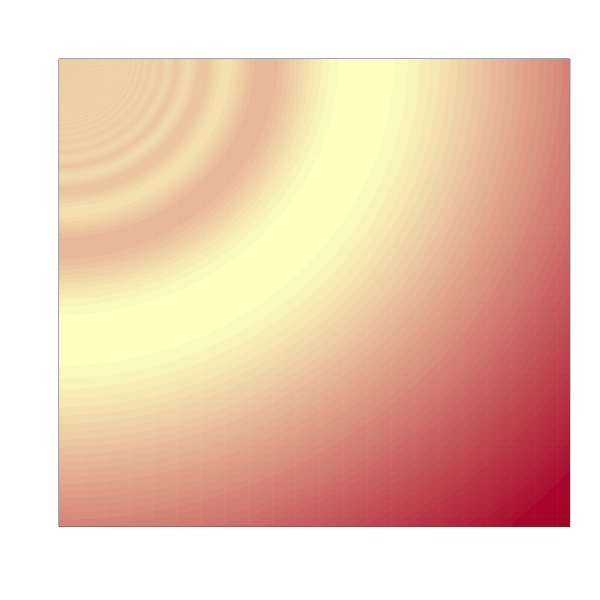}
\caption{Model 2}
\label{fig:rainbow-graphon}
\end{subfigure}
\hfill
\begin{subfigure}[t]{0.3\textwidth}
\centering
\includegraphics[width=\textwidth]{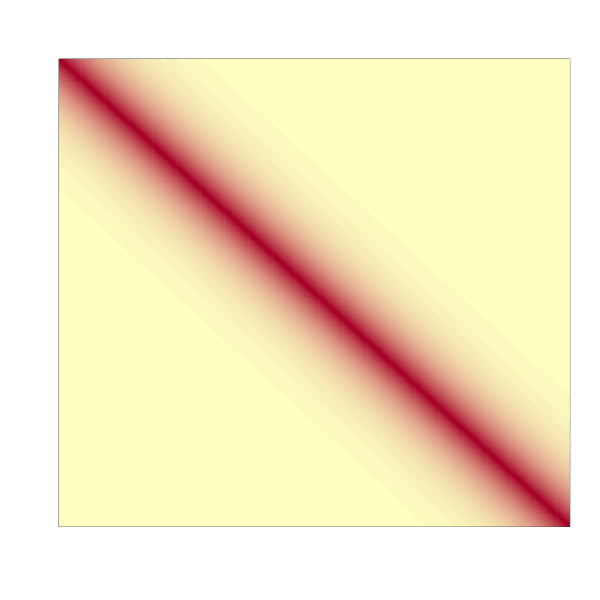}
\caption{Model 3}
\label{fig:diag-graphon}
\end{subfigure}
\caption{Three network models based on the graphon setup from \cite{zhang2015estimating}. The first model is of rank 3. The other two are full-rank models.}
\label{fig:graphons}
\end{figure}

We now evaluate the mixing methods using simulation examples. We focus on the task of estimating the network connection probability matrix, $P$. In all the examples below, we set the network size to $n=1000$, with varying density. We first evaluate our method in the graphon model settings. In particular, we use the three connection graphon connection matrices $W$ from \cite{zhang2015estimating}, for which we set $P = \alpha \cdot W$ and use $\alpha$ to control the expected average degree of the networks (Figure~\ref{fig:graphons}).

The mixing methods are based on the recommended setting discussed in Section~\ref{secsec:practical}, using SBM and DCBM fitting for $1 \le k \le K_{\max}$, and USVT. We first show that the proposed method is robust to $K_{\max}$ within a reasonable range. After that, we compare the mixing estimator with a few benchmarks. We use $K_{\max} = 15$ and $p=0.1$ as the default setting in all the benchmark comparisons. Given an estimator $\hat{P}$, performance is measured by the relative Frobenius error, defined as
$$\norm{\hat{P}-P}_F^2/\norm{P}_F^2.$$

\subsection{Comparison of mixing aggregation strategies}

\begin{figure}[H]
\centering
\begin{subfigure}[t]{0.32\textwidth}
\centering
\includegraphics[width=\textwidth]{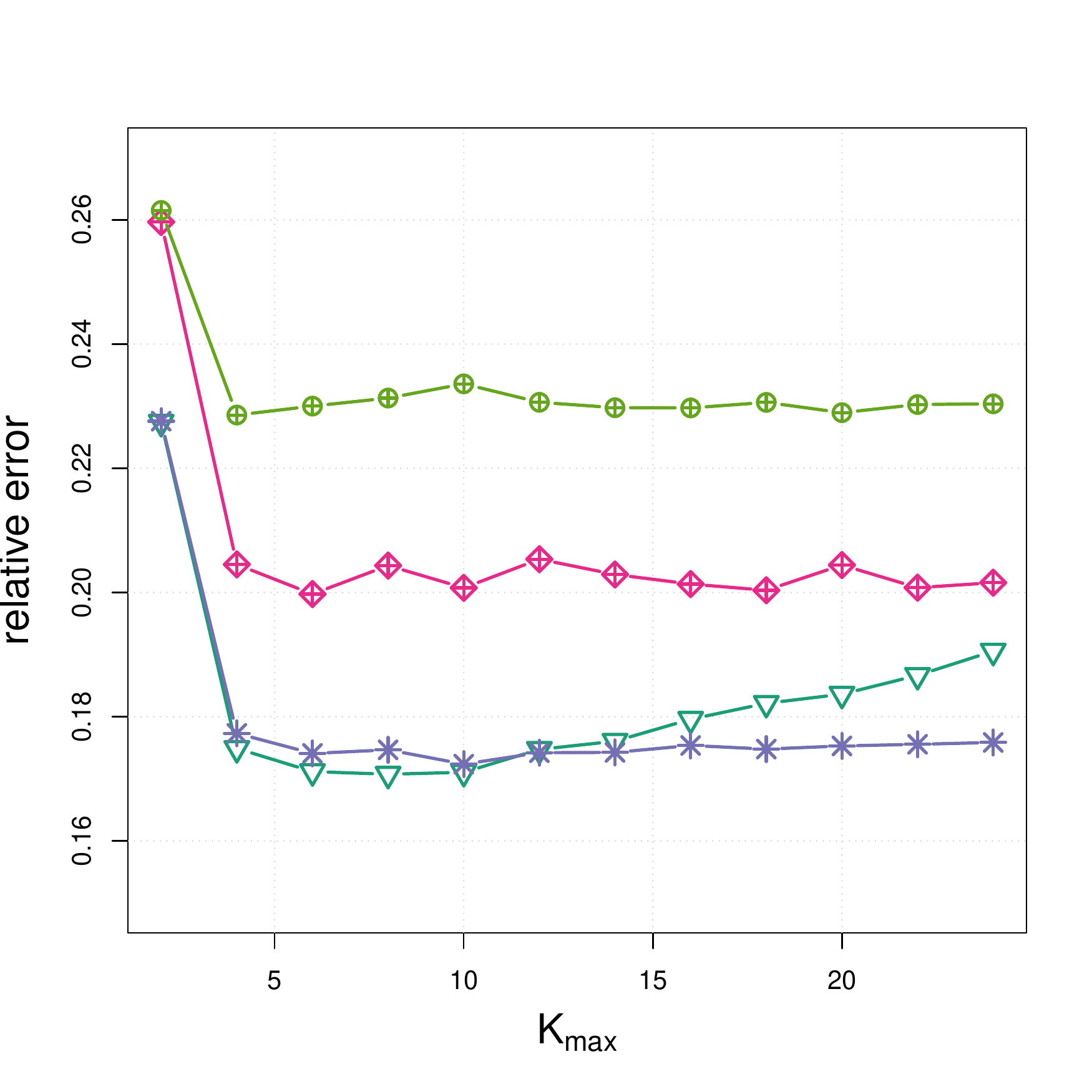}
\caption{Model 1}
\label{fig:bars-graphon-varying-K}
\end{subfigure}
\hfill
\begin{subfigure}[t]{0.32\textwidth}
\centering
\includegraphics[width=\textwidth]{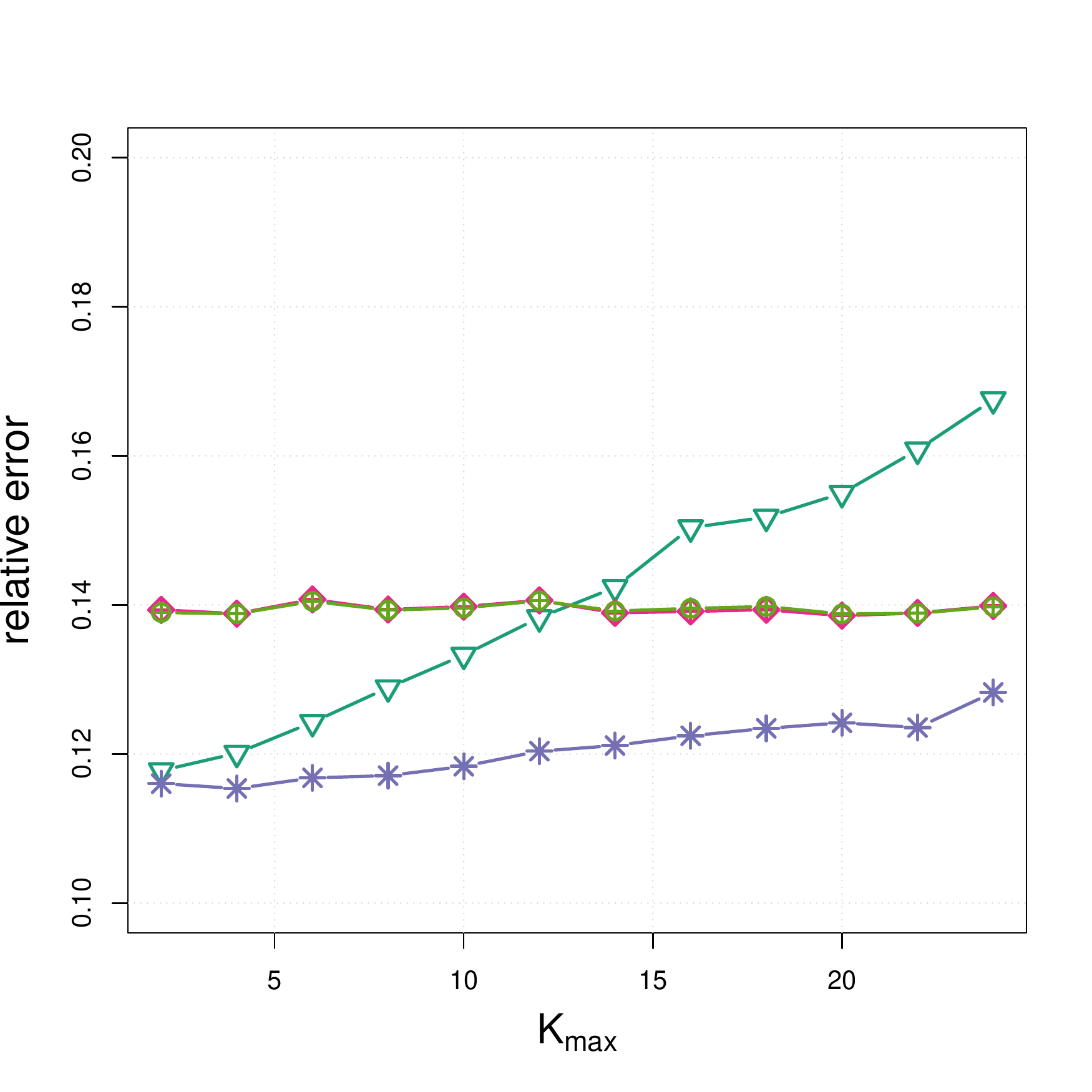}
\caption{Model 2}
\label{fig:rainbow-graphon-varying-K}
\end{subfigure}
\hfill
\begin{subfigure}[t]{0.32\textwidth}
\centering
\includegraphics[width=\textwidth]{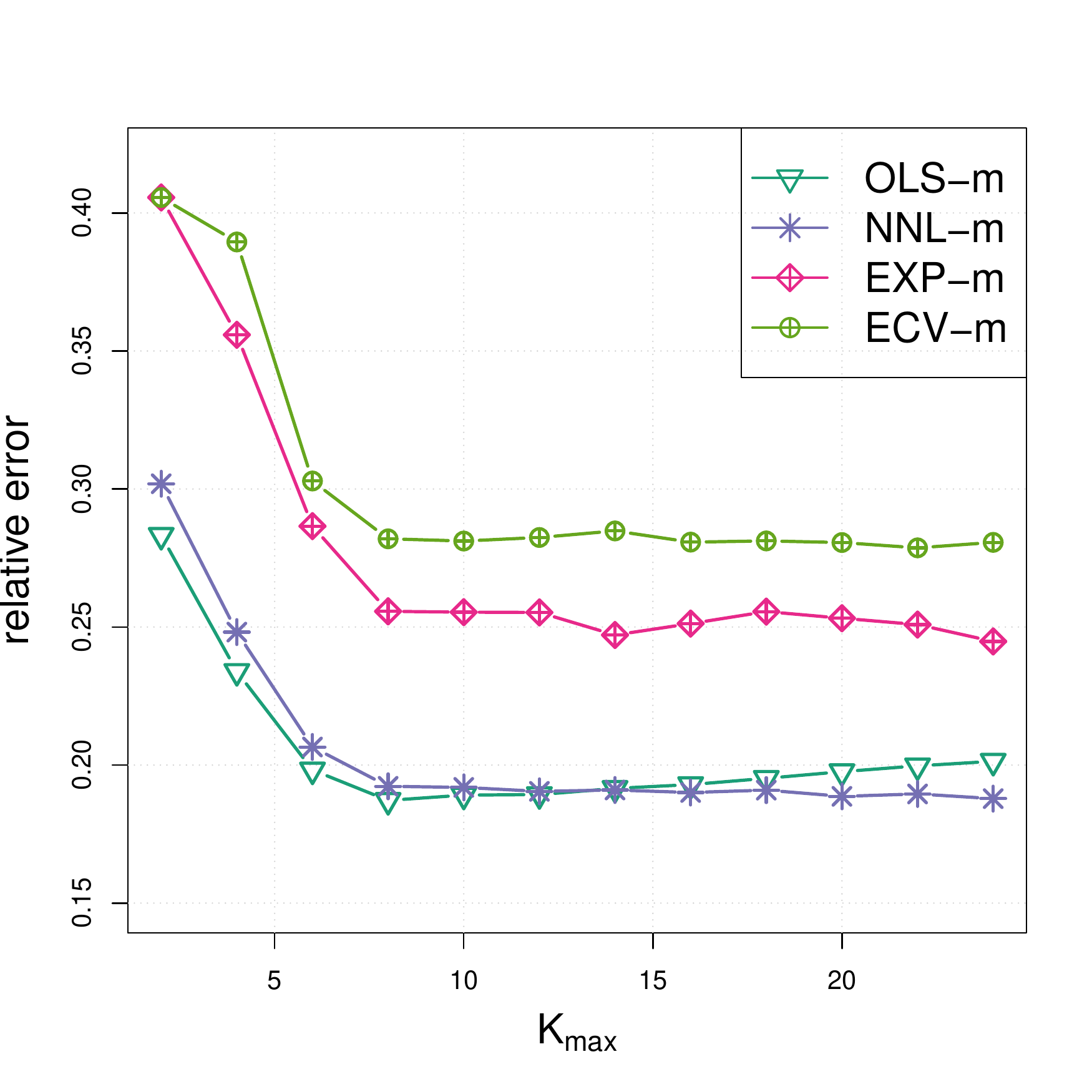}
\caption{Model 3}
\label{fig:diag-graphon-varying-K}
\end{subfigure}
\caption{Relative estimation error of $P$ with various values of $K_{\max}$ and $p=0.1$. The networks under evaluation have $n=1000$ nodes and average node degree $20$.}
\label{fig:K-value}
\end{figure}

We first want to compare different aggregation strategies for the mixing method, including linear mixing (OLS-m), NNL mixing (NNL-m), exponential mixing (EXP-m), and edge cross-validation model selection (ECV-m). Since $K_{\max}$ needs to be specified, we investigate its impact on the mixing method's performance. To demonstrate that the method is almost tuning-free, we want to show that it remains stable for a reasonable range of $K_{\max}$ values. Figure~\ref{fig:K-value} shows the estimation performance with varying $K_{\max}$ for networks with expected average node degree $20$.

As can be seen, small values of $K_{\max}$ result in underfitting and a large error. As  $K_{\max}$ increases, the error drops until overfitting kicks in. However, overall, exponential mixing, NNL mixing, and cross-validation do not suffer from overparameterization of block approximations. Linear combination, in contrast, degrades when $K_{\max}$ is large. This observation  matches our theoretical understanding of the linear estimator as we are  fitting the model with an increasing number of variables. Exponential mixing significantly outperforms cross-validation, while NNL mixing is much better than both.

\begin{figure}[H]
\centering
\begin{subfigure}[t]{0.32\textwidth}
\centering
\includegraphics[width=\textwidth]{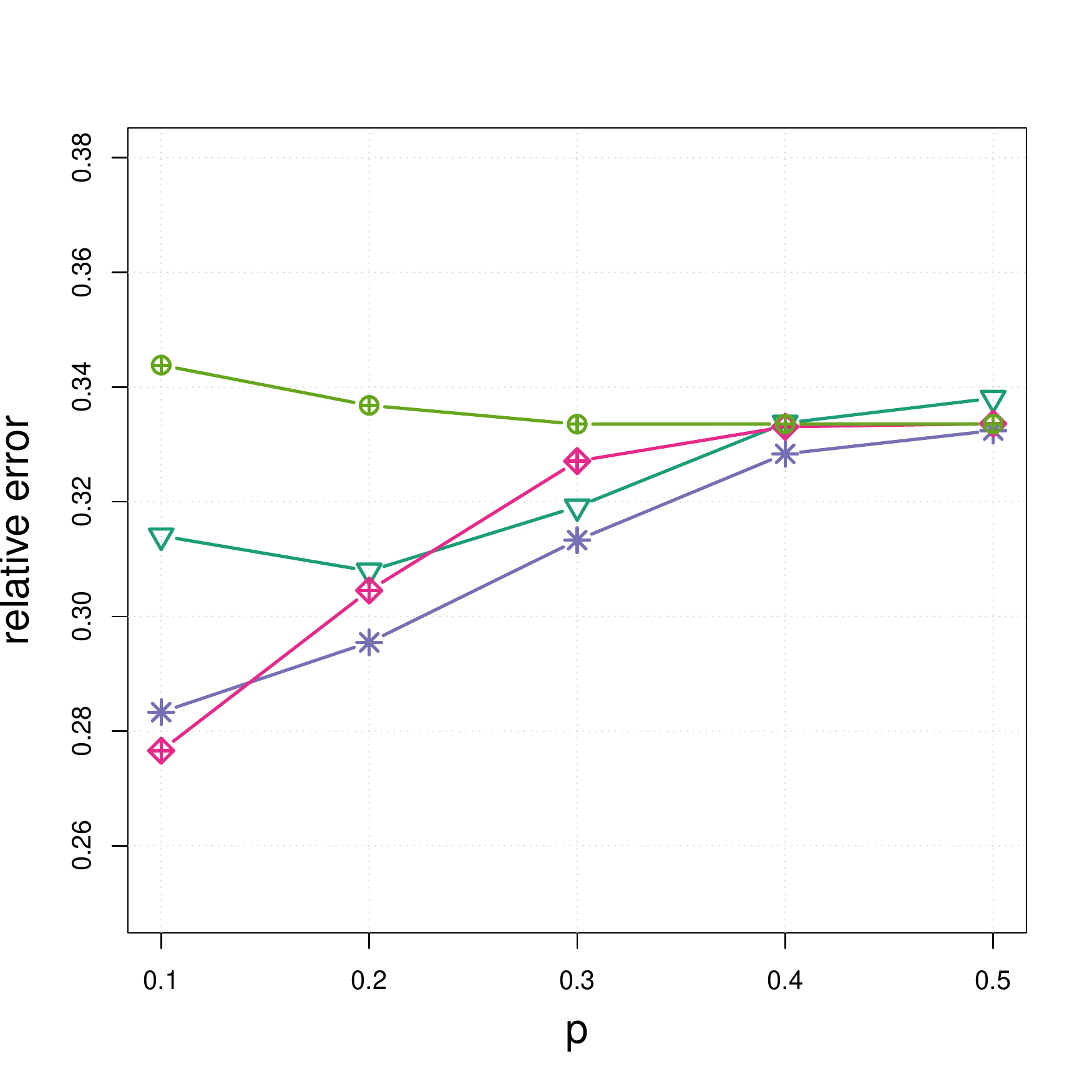}
\caption{Model 1}
\label{fig:bars-graphon-varying-rho}
\end{subfigure}
\hfill
\begin{subfigure}[t]{0.32\textwidth}
\centering
\includegraphics[width=\textwidth]{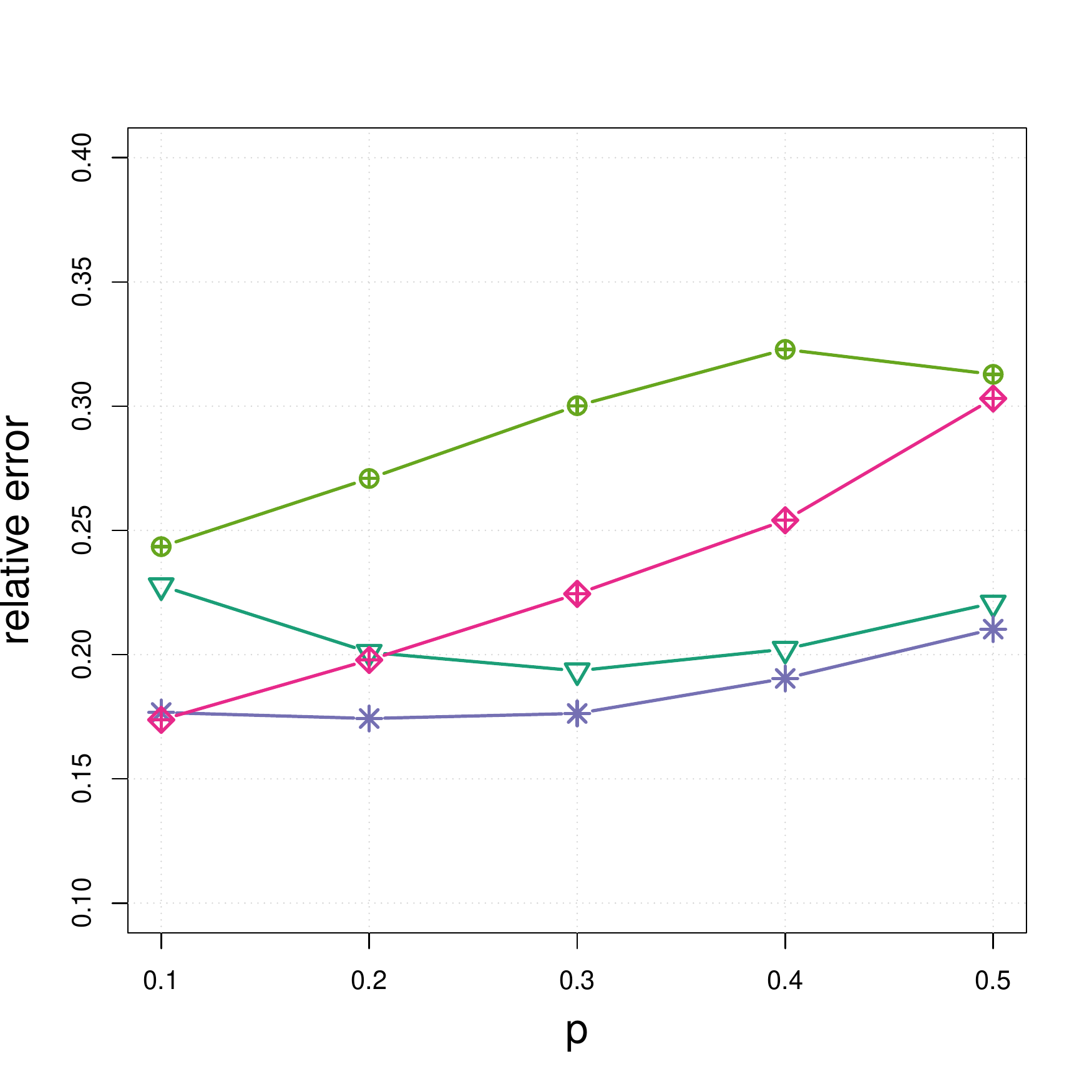}
\caption{Model 2}
\label{fig:rainbow-graphon-varying-rho}
\end{subfigure}
\hfill
\begin{subfigure}[t]{0.32\textwidth}
\centering
\includegraphics[width=\textwidth]{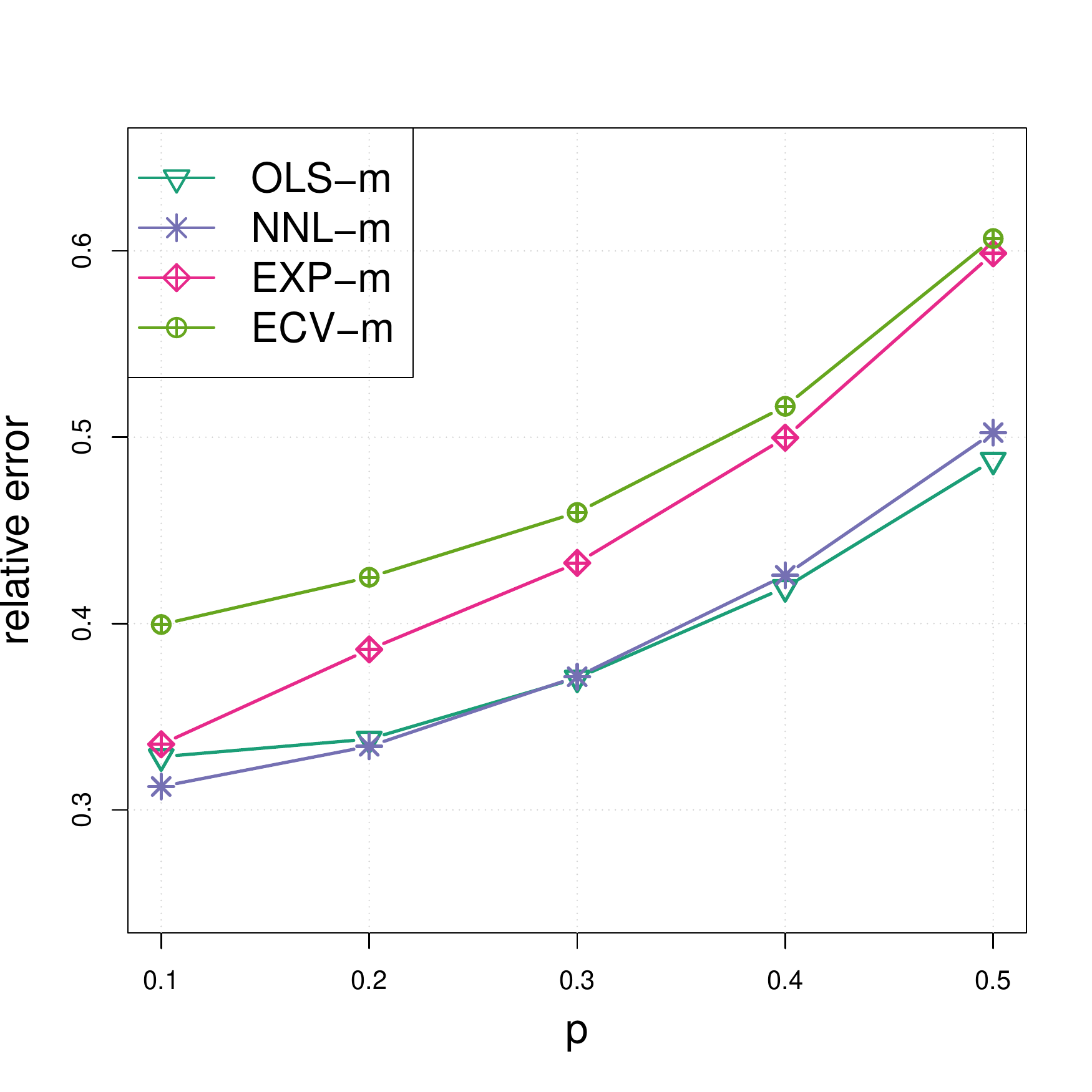}
\caption{Model 3}
\label{fig:diag-graphon-varying-rho}
\end{subfigure}
\caption{Relative estimation error of $P$ with varying $p$ and  $K_{\max} = 15$. The networks under evaluation have $m=1000$ and average node degree $20$.}
\label{fig:rho-value}
\end{figure}

Next, we evaluate the robustness of the different strategies with respect to the hold-out proportion, $p$. We evaluate their performance when $p$ varies from 0.1 to 0.5 and  $K_{\max}$ is fixed at 15 (Figure~\ref{fig:rho-value}). Overall, all the aggregation methods tend to improve as $p$ decreases within this range. This indicates that the aggregation weight determination step is easier and requires a smaller sample size than the individual model estimation step. Of all the aggregation methods, non-negative linear combination is the most robust. Linear combination also delivers competitive estimation accuracy. Based on this observation, we recommend using $K_{\max} = 15, p=0.1$ as the default configuration for network mixing estimation. 

\begin{figure}[H]
\centering
\begin{subfigure}[t]{0.32\textwidth}
\centering
\includegraphics[width=\textwidth]{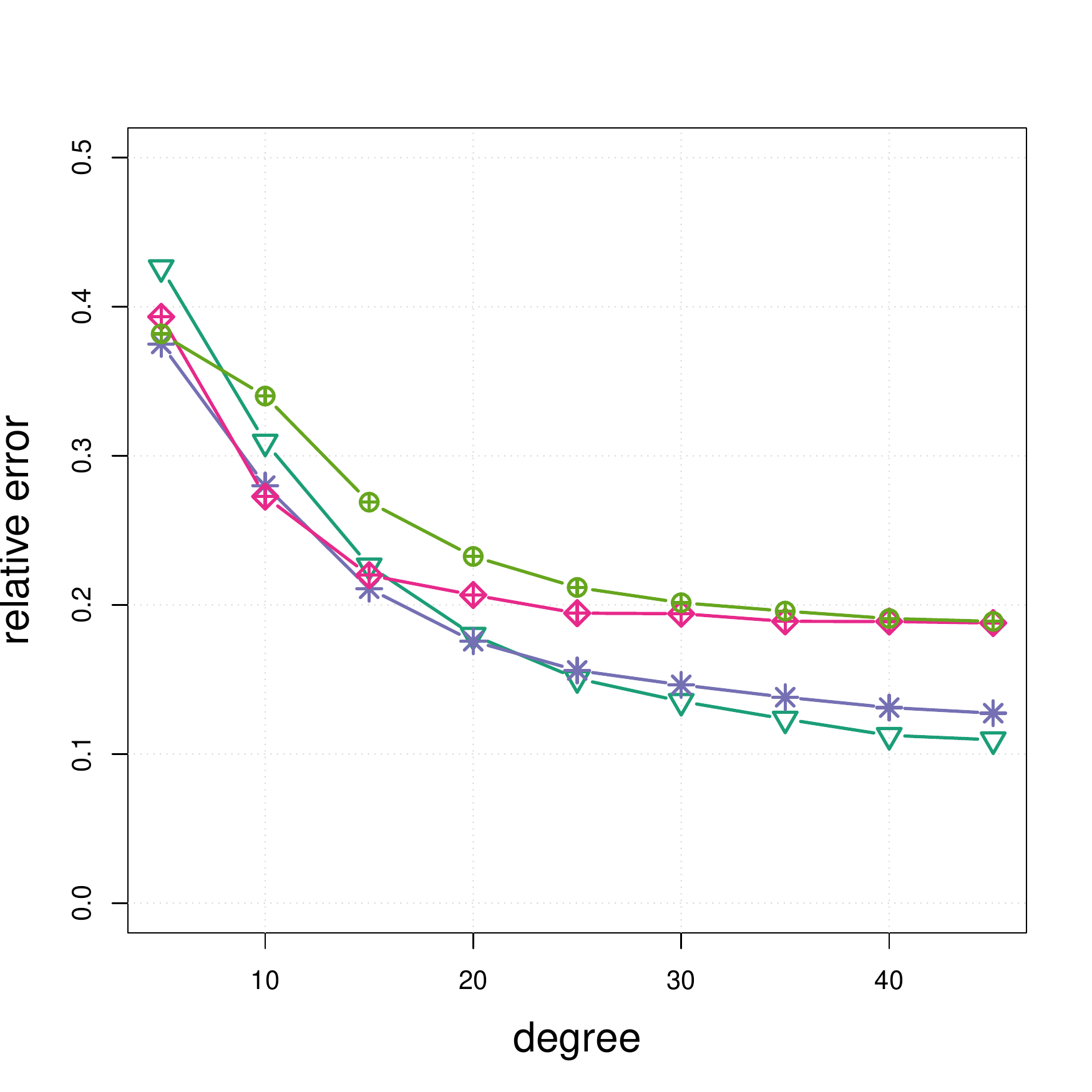}
\caption{Model 1}
\label{fig:bars-graphon-varying-d}
\end{subfigure}
\hfill
\begin{subfigure}[t]{0.32\textwidth}
\centering
\includegraphics[width=\textwidth]{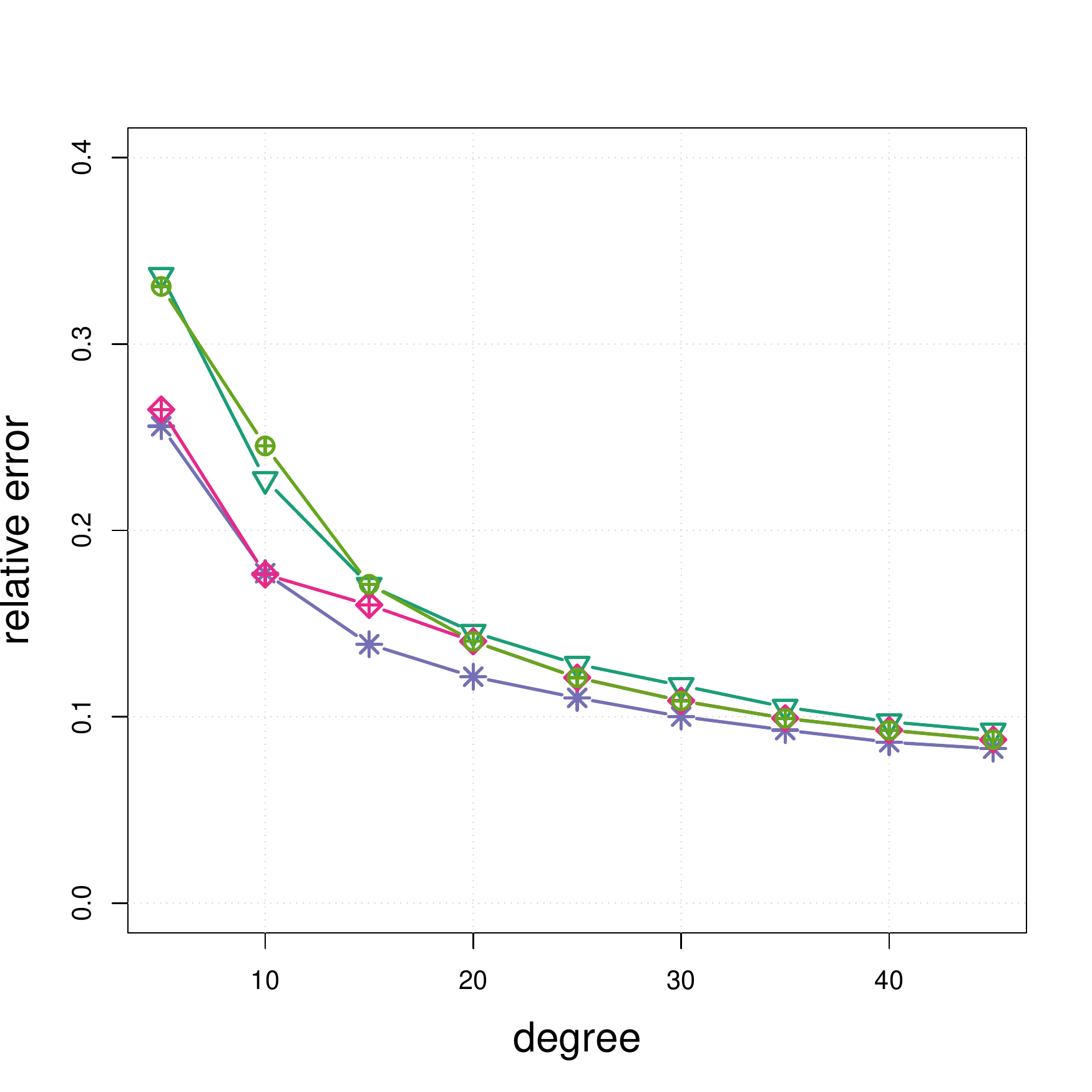}
\caption{Model 2}
\label{fig:rainbow-graphon-varying-d}
\end{subfigure}
\hfill
\begin{subfigure}[t]{0.32\textwidth}
\centering
\includegraphics[width=\textwidth]{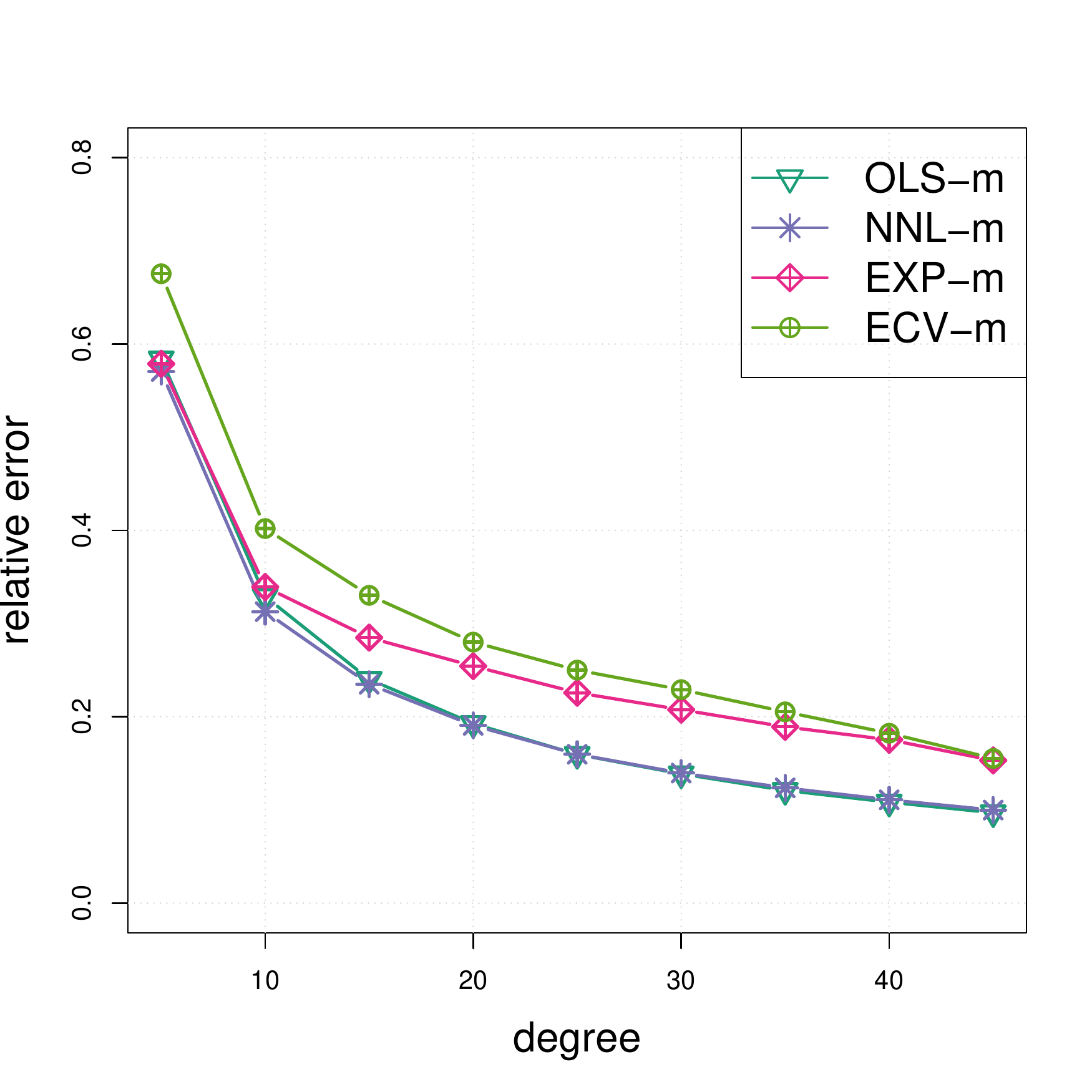}
\caption{Model 3}
\label{fig:diag-graphon-varying-d}
\end{subfigure}
\caption{Relative estimation error of $P$ with various expected average degrees. The networks under evaluation contain $n=1000$ nodes, and $K_{\max}=15, p = 0.1$ are used.}
\label{fig:d-value}
\end{figure}

Now we fix $K_{\max} = 15$ and $p=0.1$ and evaluate the estimation performance for a range of network sparsity levels,  varying the expected average degree from 5 to 45 (Figure~\ref{fig:d-value}). In the sparse regime, NNL mixing outperforms the others, while linear mixing catches up as the network becomes denser. This is predicted in Corollary~\ref{cor:comparison} as well. Exponential mixing eventually coincides with the cross-validation method, which also matches our theory.

Overall, NNL mixing is preferable compared to the other methods. Linear mixing is less robust to the choice of $K_{\max}$ and is inferior to non-negative combination in the sparse setting. However, when the network is denser, it outperforms the others. In all of the experiments to follow, we use $K_{\max}=15$ and $p=0.1$, and we believe this configuration can be used in almost all applicable tasks.

\subsection{Comparison with benchmark network estimation methods }

Now we compare the mixing method with a few benchmark network estimation methods. We split this into two parts. First, we consider the three graphon models in Figure~\ref{fig:graphons}, and we compare the mixing method to the graphon methods, which have theoretical guarantees and reasonable computational cost. These include the USVT estimator of \cite{chatterjee2015matrix}, the neighborhood smoothing (NS) method of \cite{zhang2015estimating}, and the sort-and-smooth (SAS) method of block approximation from \cite{airoldi2013stochastic} and \cite{chan2014consistent}. Second, we generate networks from two special parametric models: the SBM and the latent space model \citep{hoff2002latent}. Under these models, oracle estimations can be achieved by parametric model fitting, and they are included for comparison. The mixing, USVT, NS, and latent space model fitting \citep{ma2020universal} are all based on the R package {\em randnet} \citep{randnet}, while SAS is based on the R package {\em graphon} \citep{graphon}.

\begin{figure}[H]
\centering
\begin{subfigure}[t]{0.32\textwidth}
\centering
\includegraphics[width=\textwidth]{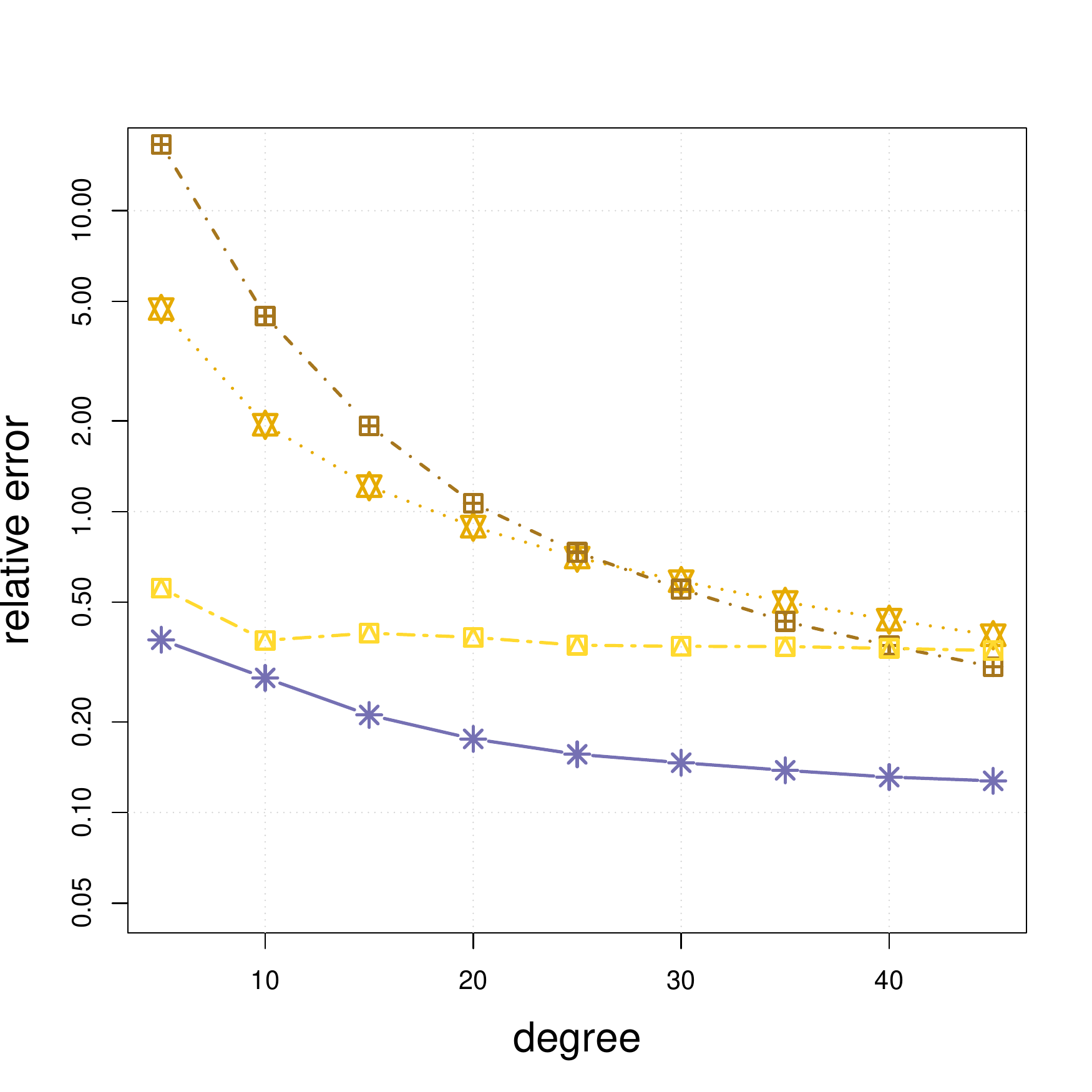}
\caption{Model 1}
\label{fig:bars-graphon-varying-density-benchmark}
\end{subfigure}
\hfill
\begin{subfigure}[t]{0.32\textwidth}
\centering
\includegraphics[width=\textwidth]{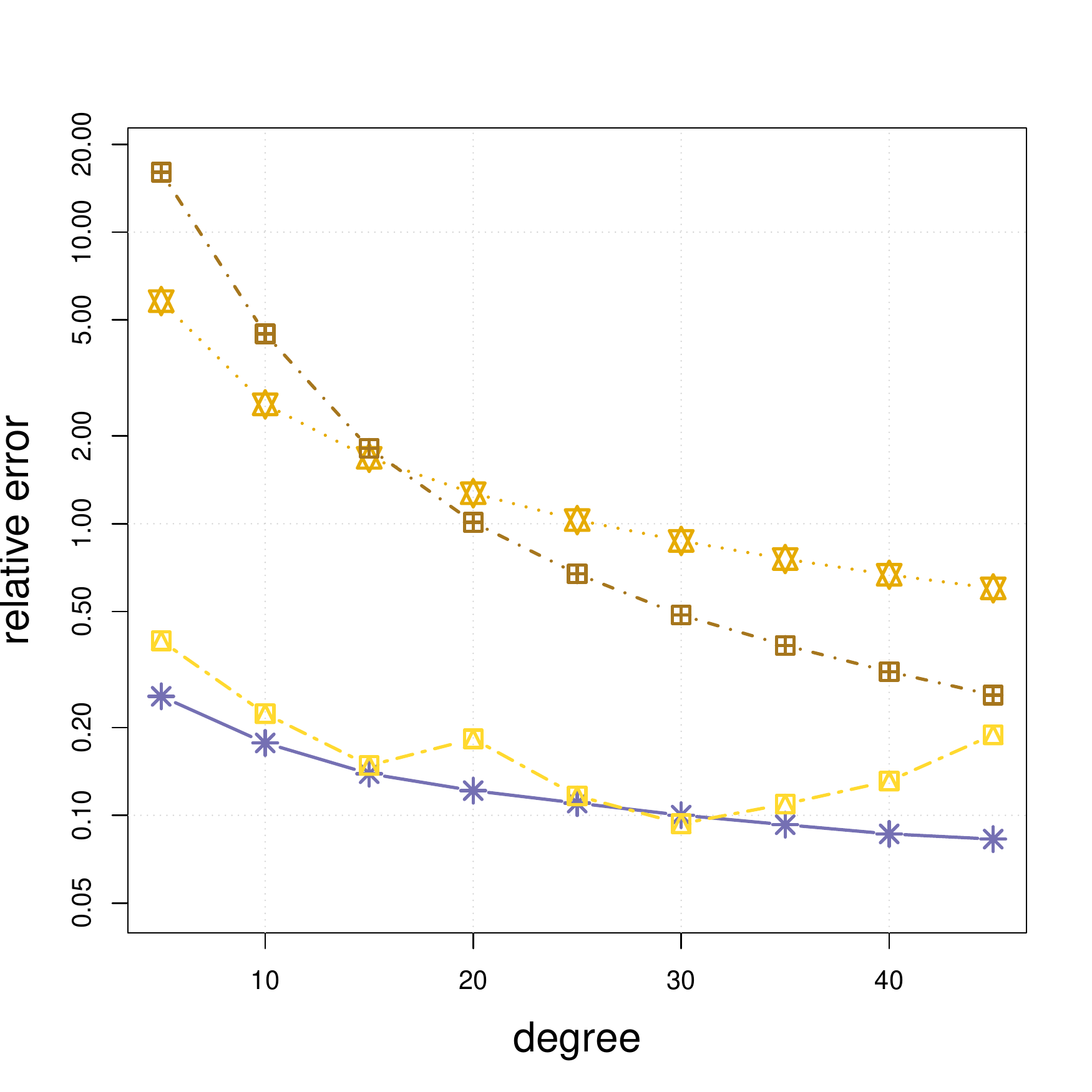}
\caption{Model 2}
\label{fig:rainbow-graphon-varying-density-benchmark}
\end{subfigure}
\hfill
\begin{subfigure}[t]{0.32\textwidth}
\centering
\includegraphics[width=\textwidth]{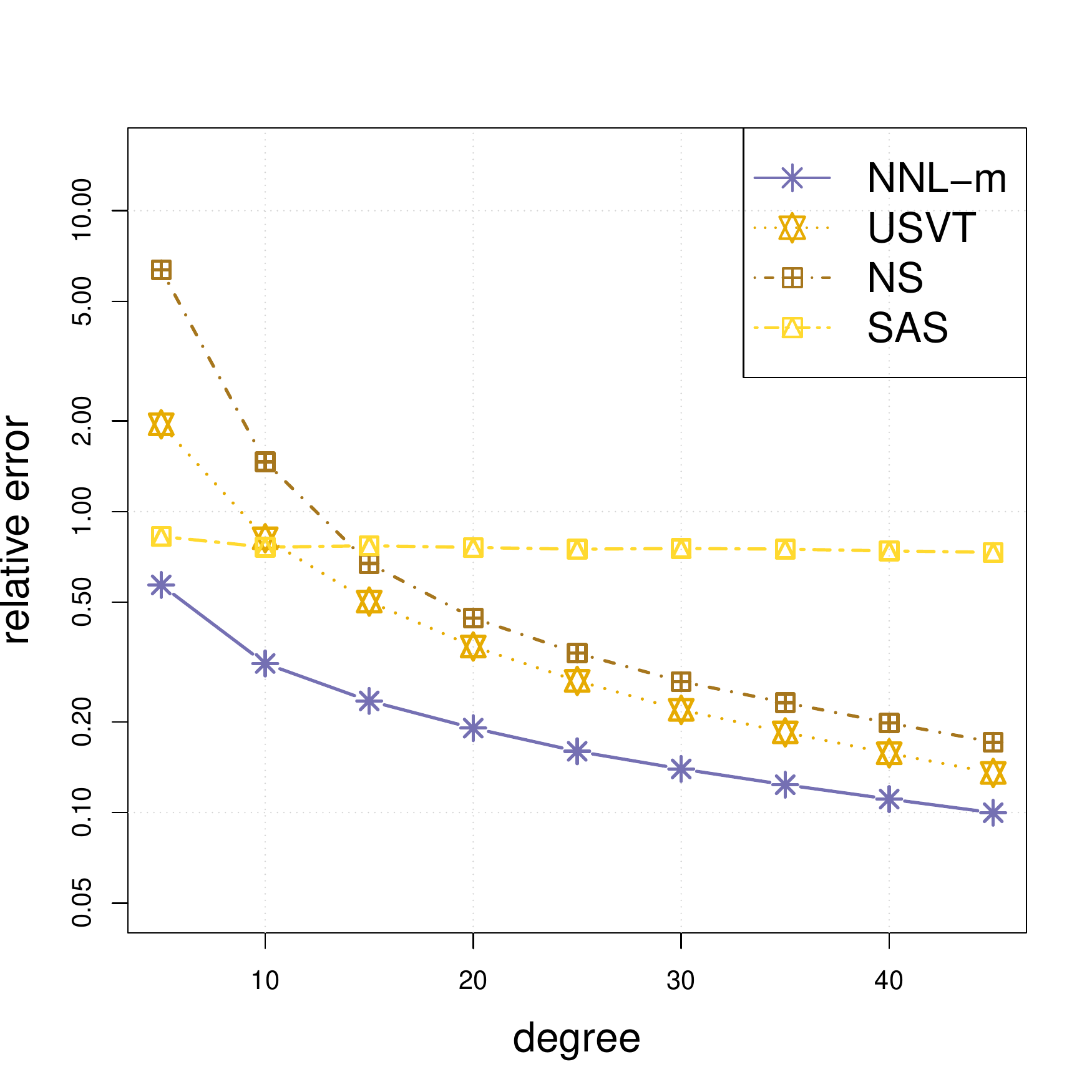}
\caption{Model 3}
\label{fig:diag-graphon-varying-density-benchmark}
\end{subfigure}
\caption{Estimation performance of the proposed mixing methods and three benchmark graphon estimation methods on synthetic networks generated from three graphon models. }
\label{fig:density-benchmark}
\end{figure}

Figure~\ref{fig:density-benchmark} shows the model estimation performance of the proposed mixing strategies and the three graphon estimation benchmarks. All four variant mixing strategies uniformly outperform the benchmark methods with all three models. The advantage of the mixing method is clear when the networks are sparse. Of the three graphon estimation methods, SAS is more accurate for the first two models, while USVT and NS are better for the third one. The difference between various mixing strategies is negligible. 

Next, we generate networks from two parametric models. The first one is the SBM, with six communities of equal size,  following the configuration of \cite{zhang2015estimating}. Under the SBM, we include two oracle estimators. Oracle1 requires the true community labels and estimates the probability matrix $P$ by averaging entries within corresponding blocks of the adjacency matrix $A$. Oracle2 knows the true SBM with six clusters but not the true community labels; it uses spectral clustering to find the node labels and averages entries within estimated blocks of $A$. In particular, notice that Oracle2 itself is automatically included as one of the individual models in the mixing. 

The second model is the latent space model (LSM), with 
$$\text{logit}(P_{ij}) = \alpha_i + \alpha_j + \langle Z_i, Z_j \rangle,$$
where $Z_i, Z_j$ are latent vectors in $\bR^4$, generated by $N(0, I_4)$ and then centralized according to \cite{ma2020universal}. Oracle1 in this case is the oracle version of the model that uses the true model structure and also the true latent dimension, estimated by  the gradient descent method of \cite{ma2020universal} with the recommended initialization. Oracle2 still assumes the correct model structure, but uses the wrong dimension (3 instead of 4), representing the possibility of dimensionality mis-specification. For reference, we also include an oracle version of the NNL mixing method, which has Oracle1 as an individual component.

\begin{figure}[H]
\centering
\begin{subfigure}[t]{0.45\textwidth}
\centering
\includegraphics[width=\textwidth]{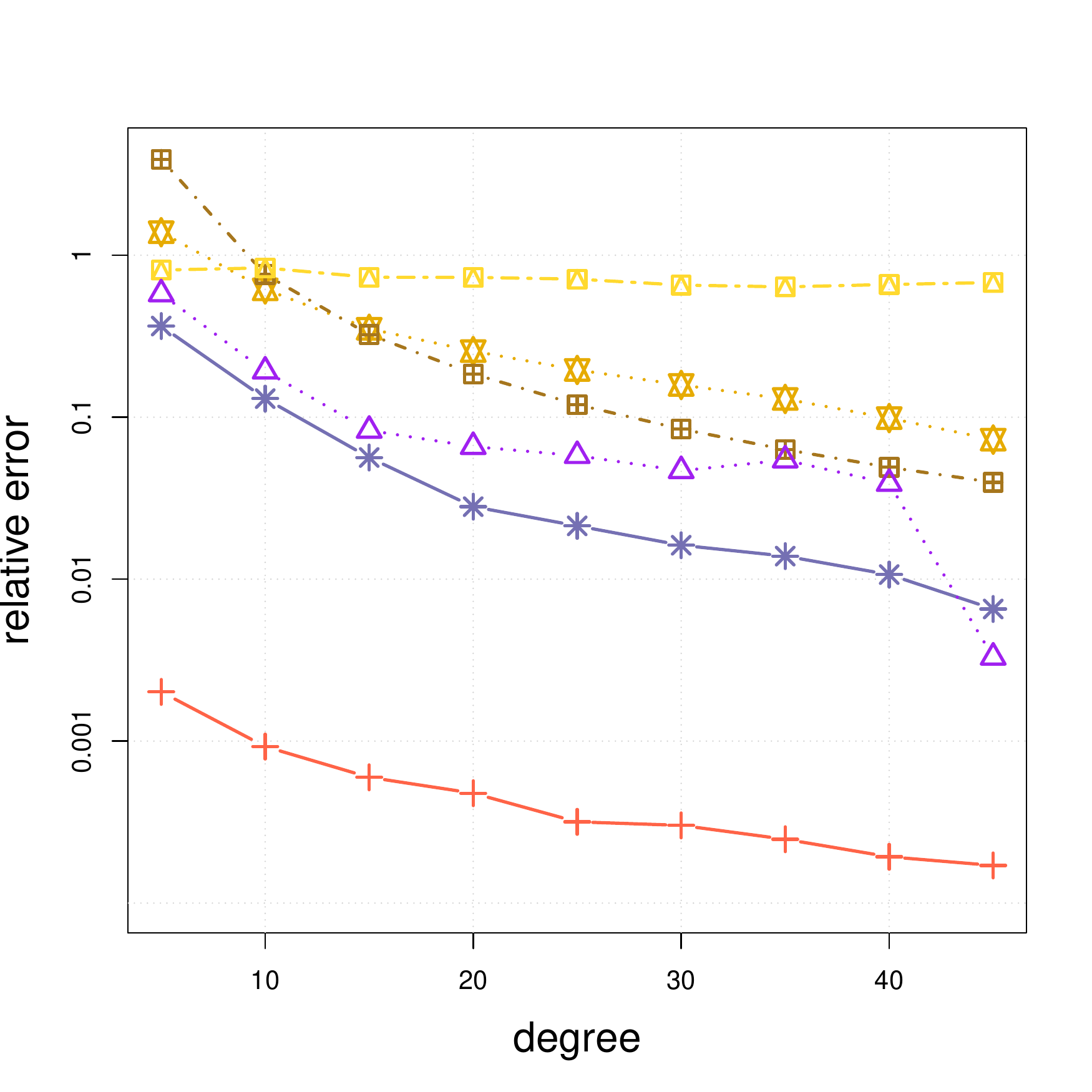}
\caption{Stochastic block model}
\label{fig:block-graphon-varying-density-benchmark}
\end{subfigure}
\hfill
\begin{subfigure}[t]{0.45\textwidth}
\centering
\includegraphics[width=\textwidth]{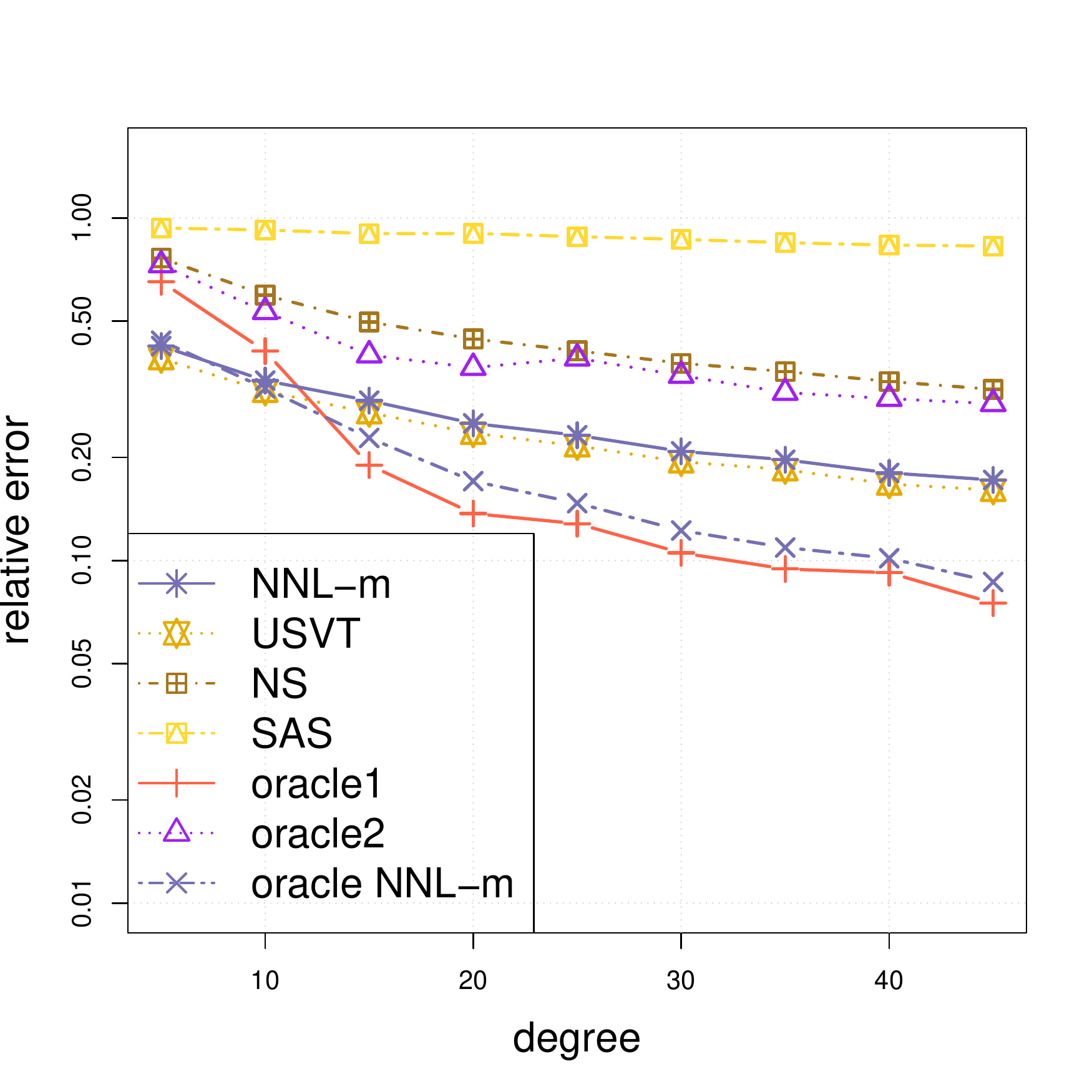}
\caption{Latent space model}
\label{fig:LSM-graphon-varying-density-benchmark}
\end{subfigure}
\caption{Comparison with benchmark graphon estimation methods on synthetic networks generated from two parametric models. }
\label{fig:parametric-benchmark}
\end{figure}

Figure~\ref{fig:parametric-benchmark} shows the performance of all the methods under the two parametric models. In the SBM setting, the mixing methods are inferior only to the unbeatable Oracle1 and are even better than Oracle2. Since the mixing procedure includes Oracle2 as an individual component, this result demonstrates the effects of ensembling multiple models. In the difficult regime, including multiple models may help stabilize the estimation and further improve the estimate of the true model when it is fitted  separately. In the LSM setting, the mixing method is again inferior only to the perfect oracle parametric estimation (Oracle1) in the dense setting and, again, it is even better than it in the sparse setting. USVT is also very effective in this setting, as indicated by the theory of \cite{chatterjee2015matrix}, and the mixing method adapts to it. NNL-m and USVT are even better than Oracle2 estimate.  The oracle NNL-m adapts to Oracle1 in the dense setting and outperforms it in the sparse setting. The comparison between the mixing method and the oracle methods under these two parametric models highlights two advantages of the mixing approach:
\begin{itemize}
 \item In the sparse regime, the estimation accuracy may be poor even if the true model is known. The mixing strategy provides a mechanism for combining simpler models to obtain a more stable estimate.
 \item In the dense regime,  the mixing approach may remain effective even if the mixing ensemble does not include the true model and can match the oracle otherwise.
\end{itemize}

%% file: data.tex
Link prediction is a widely used procedure in network science and many scientific domains to predict missing links/non-links based on a partially observed network \citep{liben2007link,lichtenwalter2010new, lu2011link}. A link prediction algorithm usually provides a set of scores corresponding to the missing links/non-links. A higher score means that it is more likely that a link exists between the corresponding nodes. When the data are generated from a random network model, the estimated edge probabilities, if available, provide natural prediction scores. One can predict that edges exist for the node pairs with scores higher than a given threshold.  \cite{ghasemian2020stacking} test a large body of link prediction algorithms on 550 real-world networks from six application domains (biological, economic, social, technological, information and transportation). They observe that no method provides superior prediction accuracy across all domains. To achieve reasonable adaptivity, they propose the optimal link prediction (OLP) strategy, which aggregates a large ensemble of link prediction algorithms using random forest. They first train random forest models and then refit the link prediction algorithms using all the training data. For this data set, OLP gives nearly optimal performance across all domains.  In this section, we explore the adaptivity of the mixing method for this task and compare it with OLP.

We first explain how to apply the mixing method to the link prediction task. In \cite{ghasemian2020stacking}, three categories of link prediction algorithms are used in the ensemble: topological methods, model-based methods, embedding methods. The linear and NNL mixing methods we have introduced can include the link prediction scores from topological/embedding algorithms as individual predictions and then use either linear combination or non-negative linear combination to aggregate them. Therefore, the mixing method in this situation is similar to OLP, except that we use linear methods instead of random forest to aggregate the available estimates. Conceptually, the advantages of the mixing method are straightforward. First, it is much more computationally efficient than OLP, especially for large networks and validation sets; this is because the sample size for random forest fitting in OLP scales as $ O(n^2)$. Second, theoretical guarantees are available for the mixing method.

The data set of all the networks and the Python implementation of OLP are provided by \cite{Ghasemian.github.olp}. \cite{ghasemian2020stacking} use 42 topological network features for link prediction. We expand this set by adding the model estimates from the canonical mixing procedure -- the SBM and DCBM estimates with $K=1,2,...,15$, and the USVT estimate. We use three different configurations of features to investigate the effectiveness of different categories of features: link prediction based on all 73 features (42 topological plus 31 model-based); link prediction based on the 42 topological features; and prediction based on the 31 model-based features only. In the model-based category, we also include the three graphon methods and the latent space model studied in Section~\ref{sec:simulation}. Notice that the model-based methods are computationally efficient, while calculating some of the 42 topological features is much more time-consuming. Therefore, the topological approaches (using all 42 features) can be computationally prohibitive for large networks.

For a stable evaluation of performance, we consider only networks with more than $200$ nodes, giving in 385 networks in total. Based on their origins, they are labeled as biological (72), economic (112), informational (10), social (108), technological (55), and transportation networks (28). Given each network, we randomly sample 10\% (capped at 20,000) of the node pairs to form the test set and use the complement subnetwork for training. This sampling scheme is different from the one used by \cite{ghasemian2020stacking}, where the test set is sampled in a balanced manner to maintain similar amounts of edges and non-edges. Although that can provide better predictive performance, it may be unrealistic in practice when one has no control over the test set. We therefore take the more natural approach to sample the test set randomly. We measure the link prediction performance of a method by the predictive area under the ROC curve (AUC) based on the test data and average it over ten independent repetitions. The predictive AUC is a widely used performance metric for the link prediction task \citep{huang2005using}.

\begin{figure}[H]
\centering
\begin{subfigure}[t]{0.48\textwidth}
\centering
\includegraphics[width=\textwidth]{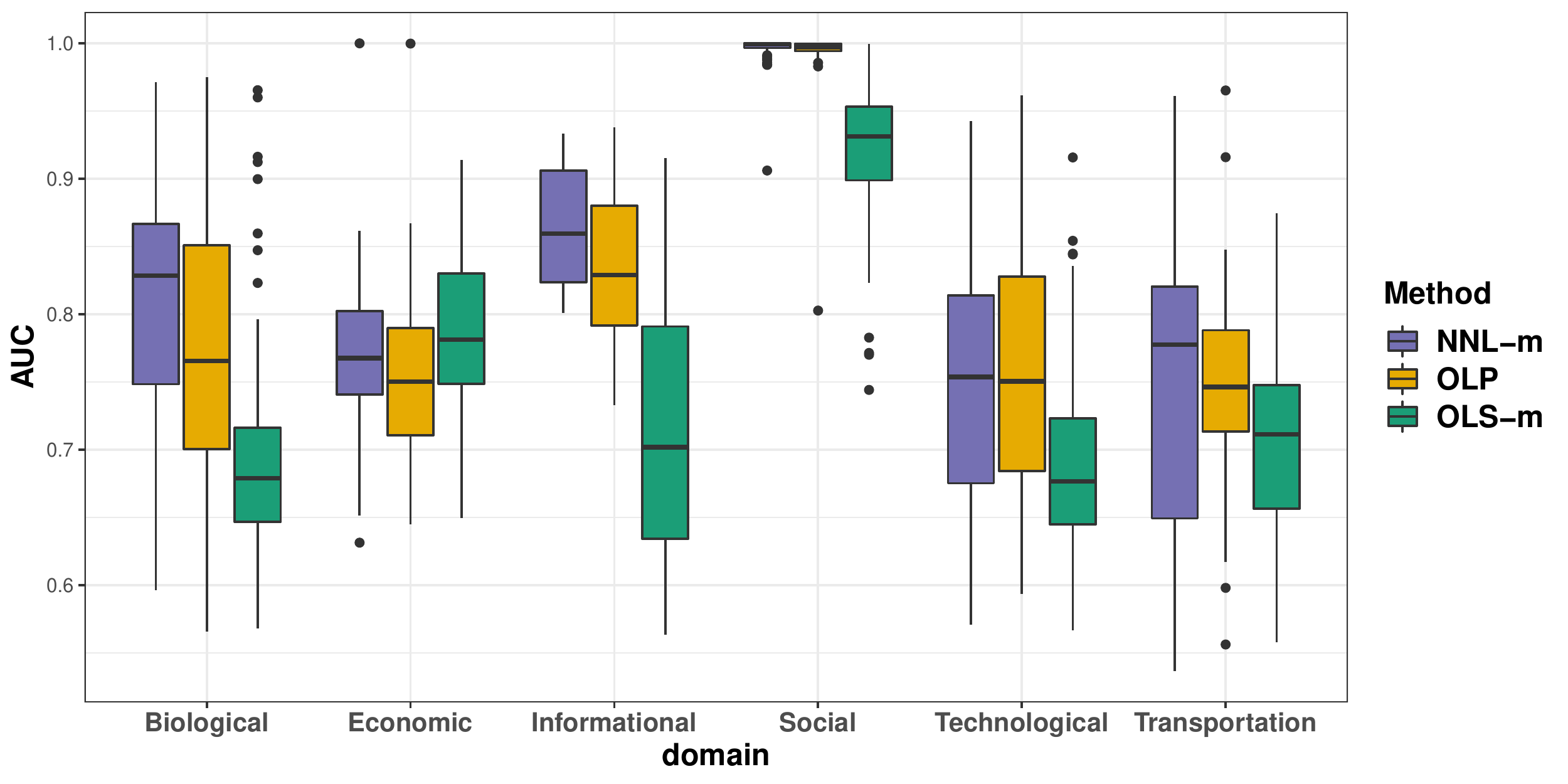}
\caption{Topological and model-based methods}
\label{fig:all-LP}
\end{subfigure}
\hfill
\begin{subfigure}[t]{0.48\textwidth}
\centering
\includegraphics[width=\textwidth]{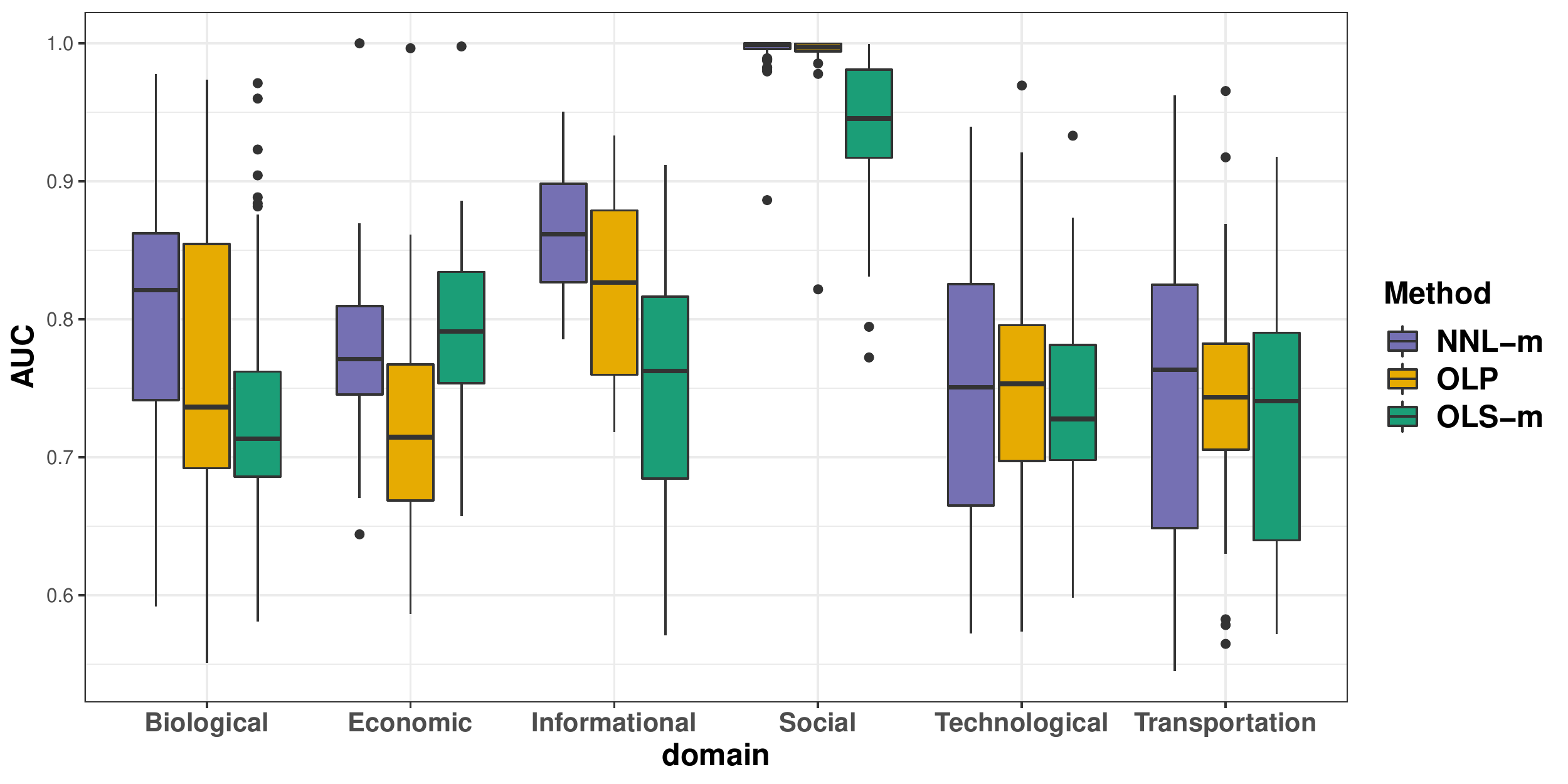}
\caption{Topological methods only}
\label{fig:topo-LP}
\end{subfigure}

\vspace{0.4cm}
\begin{subfigure}[t]{\textwidth}
\centering
\includegraphics[width=\textwidth]{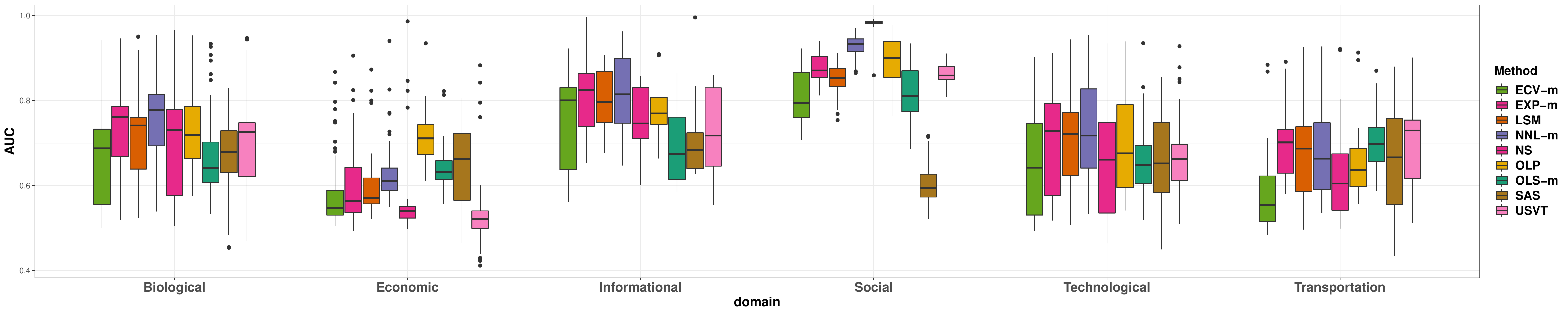}
\caption{Model-based methods only}
\label{fig:model-LP}
\end{subfigure}
\caption{Link prediction performance on 385 real-world networks from six domains.}
\label{fig:LP-evaluation}
\end{figure}

Figure~\ref{fig:LP-evaluation} shows the results for the three categories of link prediction features. The overall link prediction accuracy varies widely across domains. Social networks are easier for link prediction, while economic, technological, and transportation networks are more difficult for the same task. When topological algorithms and mode-based algorithms are implemented, the mixing method with non-negative linear combination and OLP perform similarly in five of the six categories. OLP gives slightly worse results for the sixth category, informational networks. The mixing method with linear combination is inferior to the other two in five domains but is better for economic networks. The inferiority of linear mixing may be because 73 predictors are used, and linear combination suffers from this large number of predictors. The comparison remains similar for the methods based on topological algorithms only. Overall, for model-based algorithms, the two mixing methods are still better than the other methods. The LSM and graphon estimators might be good for one category (e.g., NS for social networks) but inferior in others. The mixing methods and OLP are reasonably good in all categories. In particular, the mixing methods are still comparable to or even better than the OLP method.

\begin{figure}[H]
\centering
\begin{subfigure}[t]{0.48\textwidth}
\centering
\includegraphics[width=\textwidth]{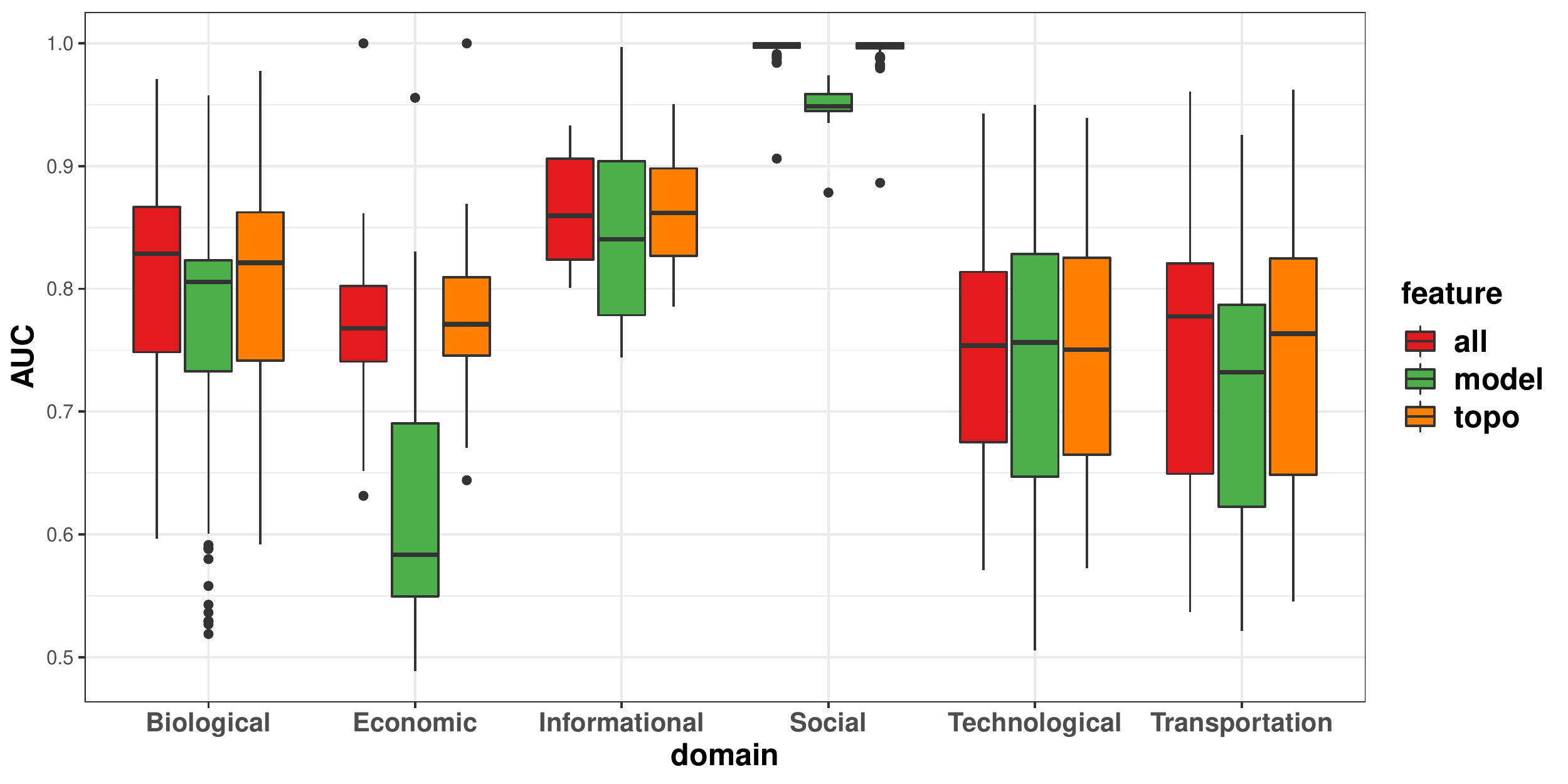}
\caption{Non-negative linear mixing}
\label{fig:feature-nnl}
\end{subfigure}
\hfill
\begin{subfigure}[t]{0.48\textwidth}
\centering
\includegraphics[width=\textwidth]{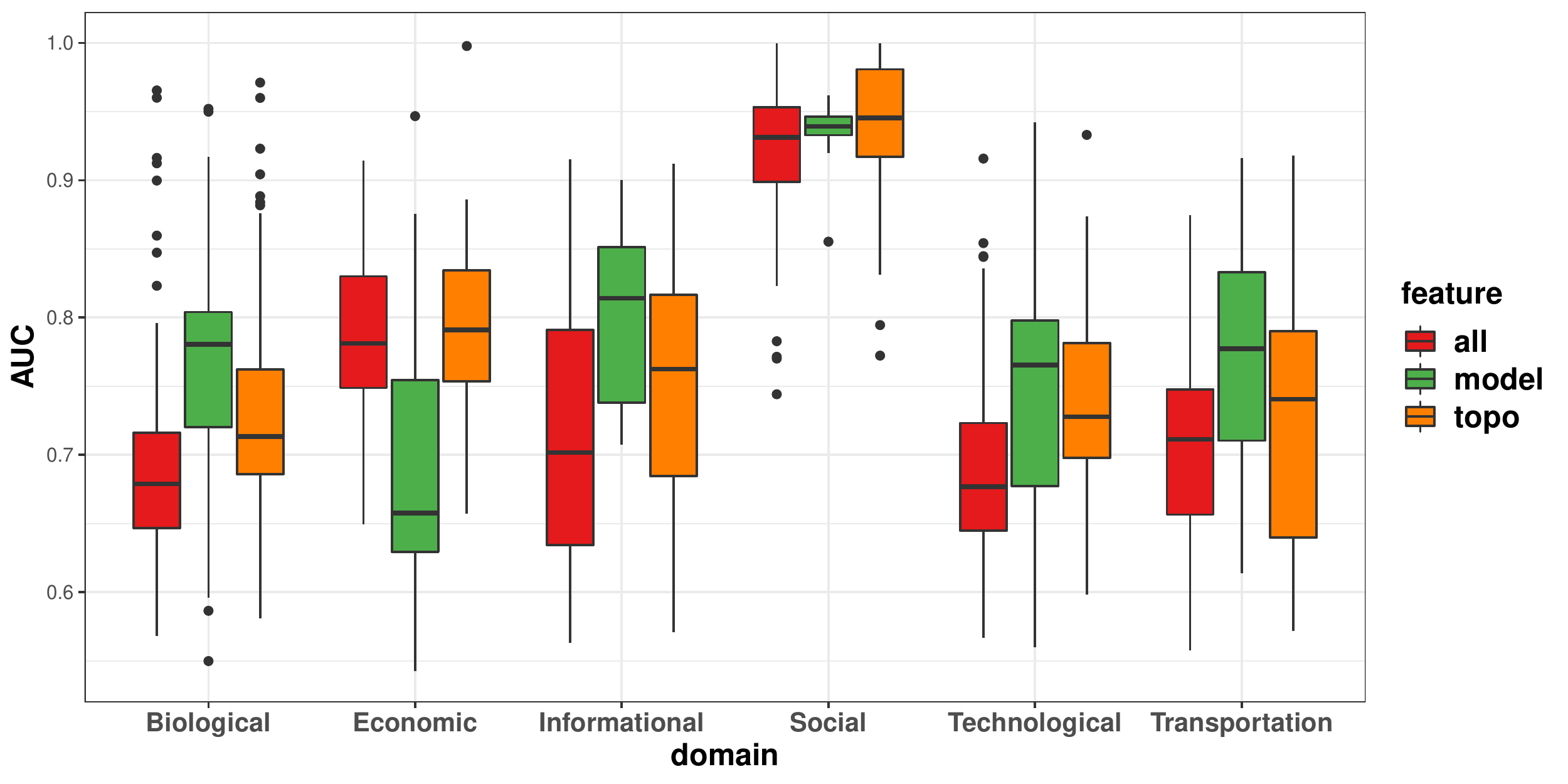}
\caption{Linear mixing}
\label{fig:feature-linear}
\end{subfigure}

\vspace{0.4cm}

\begin{subfigure}[t]{0.48\textwidth}
\centering
\includegraphics[width=\textwidth]{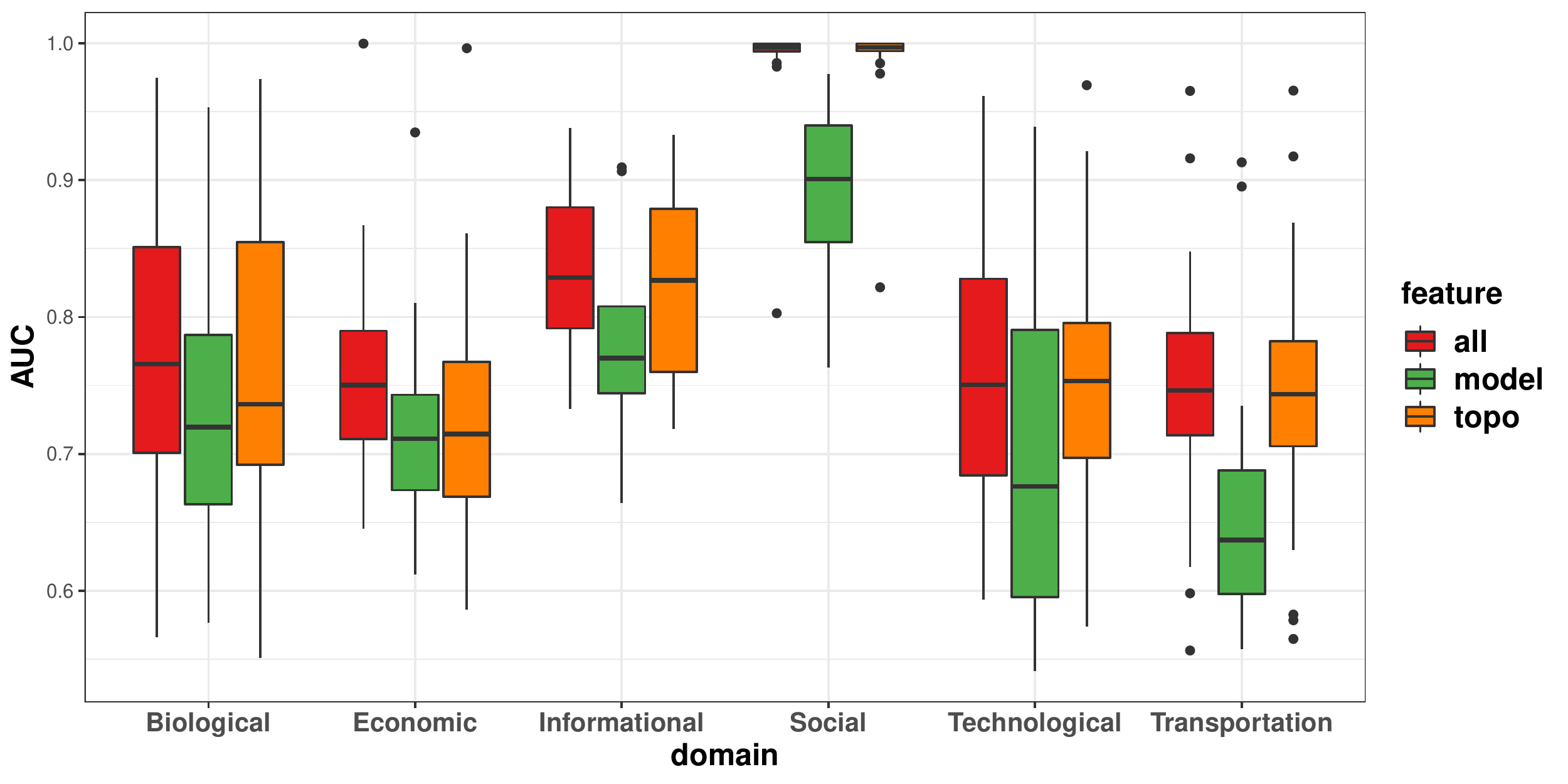}
\caption{OLP}
\label{fig:feature-OLP}
\end{subfigure}

\caption{Mixing performance using different features.}
\label{fig:LP-feature}
\end{figure}

Figure~\ref{fig:LP-feature} compares the performance of each mixing strategy for different categories of features. Both OLP and non-negative linear mixing show the same pattern: the model-based prediction is worse than the topology-based prediction, while the version using both the model and topological features has similar performance as the topology-based algorithms. Linear mixing using all features is never superior in all domains. This is likely because it becomes unstable when many mixing features are used.

As practical guidance, we also want to briefly mention the speed of the above (model-based) methods. USVT is only based on the SVD of the adjacency matrix, which is usually very sparse. It is usually the most efficient one. The mixing methods need additional modeling fitting after SVD (note that the mixing needs only one round of SVD as well), but that additional cost is relatively small. So with an efficient SVD implementation \citep{irlba,RSpectra}, SVD and USVT can easily handle networks of size $O(10^5)$ on a single laptop. In principle, SAS has a similar speed, though we observe that it is slightly slower than mixing. OLP can generate all the features in the same way as the mixing methods, but random forest fitting can be much slower, and model-fitting cost becomes the major bottleneck. In our experiments, while OLP is feasible for networks with a few thousand nodes, it is much slower than mixing. The gradient method for the latent space model and neighborhood smoothing can handle networks of moderate size but may become too slow if the size is larger than 3000.

In summary, though the OLP method is claimed to be nearly optimal for link prediction, we find that the NNL mixing strategy delivers comparable or slightly better accuracy and adaptivity across all domains. Given its additional theoretical guarantees and much higher computational efficiency, we believe that the mixing method can generally serve as an off-the-shelf link prediction algorithm.

%% file: discussion.tex
We have introduced a mixing strategy for network estimation. It can be used as an off-the-shelf method for network modeling with both flexibility and scalability. The method is designed according to geometrical insights into the network modeling problems, and we have shown the advantage of the design theoretically. We have also demonstrated its competitive performance empirically in link prediction problems.

The current study could be extended in several directions for future follow-up. In this paper, we focus on the accuracy of the mixing estimator for a network model. In many network analysis problems, the crucial quantities can differ from the properties of the network model $P$ itself. For example, in regression inference of network-linked data \citep{le2020linear}, the recovery of the network projection operator is critical. Intuitively, using the mixing estimator should render a more robust inference. The theory of this type of inference warrants further study. Another direction is to extend the mixing strategy to dynamic network modeling problems \citep{kim2018review}. Flexible dynamic network models tend to be computationally challenging to fit. Given its flexibility and computational efficiency, it would be interesting to see whether the mixing method can help alleviate this difficulty.

%% file: Proof.tex
\section{Exponential weight mixing}\label{sec: exp weights}

\begin{proof}[Proof of Theorem~\ref{thm:exp weights}]
From \eqref{eq:exp weights}, we have 
\begin{eqnarray*}
\pi_r&\propto&\exp\left(-\sum_{(i,j)\in\Omega} \left(A_{ij}-\hat{P}^{(r)}_{ij}\right)^2\right)\\
&\propto&\exp\left(-\sum_{(i,j)\in\Omega} \left(A_{ij}-\hat{P}^{(r)}_{ij}\right)^2+\sum_{(i,j)\in\Omega} (A_{ij}-P_{ij})^2\right) =: \exp\left(-S^{(r)}\right),
\end{eqnarray*}
where $\propto$ denotes ``proportional to''.
%For convenience, instead of comparing these weights directly, we first multiply them by a random factor not depending on $j$ (the key point), and then compare the resulting products 
%\begin{eqnarray*}
%\exp\left(-\sum_{(u,v)\in\Omega^c} (A_{uv}-P^j_{uv})^2+\sum_{(u,v)\in\Omega^c} (A_{uv}-P_{uv})^2\right) =: \exp(-S^j).
%\end{eqnarray*}
Rewrite $S^{(r)}$ as follows:
\begin{eqnarray*}
S^{(r)} = \sum_{(i,j)\in\Omega}\left(P_{ij}-\hat{P}^{(r)}_{ij}\right)\left(2A_{ij}-P_{ij}-\hat{P}^{(r)}_{ij}\right)=:\sum_{(i,j)\in\Omega} X_{ij}^{(r)}.
\end{eqnarray*}
Denote $\rho = \max_{i,j}P_{ij}$. 
Conditioning on $A(\Omega^c)$ (thus on $\hat{P}^{(r)}$), $S^{(r)}$ is the sum of independent random variables $X_{ij}^{(r)}$ with mean and variance given by
\begin{eqnarray*}
\e X_{ij}^{(r)} = \left(P_{ij}-\hat{P}^{(r)}_{ij}\right)^2, \quad 
\text{var}\left(X_{ij}^{(r)}\right) = 4P_{ij}(1-P_{ij})\left(P_{ij}-\hat{P}^{(r)}_{ij}\right)^2 \le 4\rho\left(P_{ij}-\hat{P}^{(r)}_{ij}\right)^2.
\end{eqnarray*}
Since $\big|X_{ij}^{(r)}\big|\le 8$, by Bernstein's inequality, 
\begin{eqnarray*}
\p\left(\big|S^{(r)}-\e S^{(r)}\big|>t\right)\le 2\exp\left(\frac{-t^2/2}{4\rho\displaystyle \sum_{(i,j)\in\Omega}\left(P_{ij}-\hat{P}^{(r)}_{ij}\right)^2+8t/3}\right).
\end{eqnarray*}
%If the following condition holds
%\begin{eqnarray*}
%\log (k+n) \le \frac{36}{\max_{(u,v)\in\Omega^c}(P_{uv}-P_{uv}^j)^2}\sum_{(u,v)\in\Omega^c} P_{uv}(1-P_{uv})(P_{uv}-P^j_{uv})^2 
%\end{eqnarray*}
Since $m = o(n^2)$, we have
\begin{eqnarray}\label{eq:concentration S}
\Big|S^{(r)}-\sum_{(i,j)\in\Omega}\left(P_{ij}-\hat{P}_{ij}^{(r)}\right)^2\Big| \le  C\left(\rho\log n \sum_{(i,j)\in\Omega} (P_{ij}-\hat{P}^{(r)}_{ij})^2\right)^{1/2}+C\log n
\end{eqnarray}
for all $1\le r \le m$ with high probability  for a sufficiently large constant $C$. In other words, $S^{(r)}$ concentrates well around the mean $\sum_{(i,j)\in\Omega}\big(P_{ij}-P_{ij}^{(r)}\big)^2$, and the  mixture of $P^{(r)}$ puts exponentially large weight on the estimator $P^{(r)}$ with the smallest error $\|P^{(r)}(\Omega)-P(\Omega)\|_F^2$. 

\medskip
Let $r^* = \argmin_{1\le r \le m}\|P^{(r)}(\Omega)-P(\Omega)\|_F^2$ and denote 
$$x_r=\|P^{(r)}(\Omega)-P(\Omega)\|_F, \quad x=\|P^{(r^*)}(\Omega)-P(\Omega)\|_F.
$$
Consider the index set  
$$
I=\left\{r\in [m]:x_r\le x+2C\sqrt{\rho\log n}+ \frac{2C\log(nm)}{x}\right\}.
$$
By the triangle inequality,
\begin{eqnarray*}
\Big\|\sum_{r=1}^m \pi_r P^{(r)}(\Omega)-P(\Omega)\Big\|_F \le \sum_{r=1}^m \pi_r \|P^{(r)}(\Omega)-P(\Omega)\|_F 
= \sum_{r\in I} \pi_r x_r + \sum_{r\not\in I} \pi_r x_r.
\end{eqnarray*} 
From the definition of $I$ we get 
\begin{eqnarray*}
\sum_{r\in I} \pi_r x_r \le x+2C\sqrt{\rho\log n}+\frac{2C\log(nm)}{x}.
\end{eqnarray*}
For the second sum, consider $r\not\in I$. Recall that $\pi_r\propto \exp(-S^{(r)})$ and by \eqref{eq:concentration S}, $S^{(r)}$ and $S^{(r^*)}$ concentrate around $x_r^2$ and $x^2$, respectively. Therefore with high probability we have
\begin{eqnarray*}
\pi_r = \frac{\exp(-S^{(r)})}{\sum_r \exp(-S^{(r)})} &\le& \exp\left(S^{(r*)}-S^{(r)}\right) \le \exp\left(x^2-x_r^2+2C\sqrt{\rho\log n}(x+x_r)+2C\log n\right).
\end{eqnarray*} 
Since $r\not\in I$, 
\begin{eqnarray*}
x_r^2-x^2-2C\sqrt{\rho\log n}(x_r+x) = (x_r+x)(x_r-x-2C\sqrt{\rho\log n}) \ge 4C\log(nm),
\end{eqnarray*}
and consequently,  
\begin{eqnarray*}
\pi_r \le \exp\left(-2C\log(nm)\right)
= \left(nm\right)^{-2C}.
\end{eqnarray*}
Since $x_r\le n$, by choosing $C>1$, we get
$$
\sum_{r\not\in I} \pi_r \|P^{(r)}(\Omega)-P(\Omega)\|_F \le \sum_{r\not\in I} n(nm)^{-2C} = o(1). 
$$
In summary, we have proved that with high probability,  
\begin{eqnarray*}
\Big\|\sum_{r=1}^m \pi_r P^{(r)}(\Omega)-P(\Omega)\Big\|_F \le \min_{1\le r\le m}\big\|P^{(r)}(\Omega)-P(\Omega)\big\|_F+ \delta,
\end{eqnarray*}
where  
\begin{eqnarray*}
\delta = 3C\Big(\log n\cdot\max_{i,j}P_{i,j}\Big)^{1/2}+\frac{3C\log(nm)}{\min_{1\le r\le m}\|P^{(r)}(\Omega)-P(\Omega)\|_F}.
\end{eqnarray*}
In case $\min_{1\le r\le m}\|P^{(r)}(\Omega)-P(\Omega)\|_F\le 1$, we can replace $\log(nm)/x$ in  the definition of $I$  by $\log^{1/2}(nm)$ and repeat the above argument. Again, for constant $C>1$, the same derivations can still go through and the resulting error would be 
$$\delta = 3C\Big(\log n\cdot\max_{i,j}P_{i,j}\Big)^{1/2}+3C\log^{1/2}(nm).
$$
Consequently, the error can be improved as follows:
\begin{eqnarray*}
\delta = 3C\Big(\log n\cdot\max_{i,j}P_{ij}\Big)^{1/2}+3C\min\left\{\frac{\log(nm)}{\min_{1\le r\le m}\|P^{(r)}(\Omega)-P(\Omega)\|_F}, \ \log^{1/2}(nm)\right\}.
\end{eqnarray*}
Rewriting $3C$ as a new constant $C$ completes the proof.
\end{proof}

\section{Non-negative linear aggregation}

%Let $Y=\mu+\epsilon\in\mathbb{R}^N$ where $\epsilon$ is a vector of centered Bernoulli random variables with success probabilities $\mu$. In the network setting, $N\approx n^2$, where $n$ is the number of nodes in the network.
%Assume that $\theta_1,...,\theta_m$ are sufficiently correlated estimates of $\mu$ in the sense that
%\begin{eqnarray}\label{eq:cor assumption}
%\min_{1\le i,j\le m}\frac{\langle \theta_i,\theta_j\rangle}{\|\theta_i\|\cdot\|\theta_j\|} \ge 1-\delta
%\end{eqnarray}
%for some $\delta\in[0,1]$. Denote by $P_L$ and $P_C$ the linear projection onto the subspace spanned by $\theta_1,...,\theta_m$ and the projection onto the cone $\mathcal{C}$ generated by $\theta_1,...,\theta_m$, respectively. Note that both the linear and non-negative linear estimates of $\mu$ incur the error $\|\mu-P_L\mu\|$, so for the comparison purpose, we will focus on the errors of the estimates with respect to $P_L\mu$. 

\begin{proof}[Proof of Theorem~\ref{thm:nnls}]
Denote $E = A(\Omega)-P(\Omega)$.  By the triangle inequality,
%\begin{eqnarray*}
%\|\Pi_\mathcal{C} A(\Omega)-P(\Omega)\|_F &=& \|\Pi_\mathcal{C}E + \Pi_\mathcal{C}P(\Omega)-P(\Omega)\|_F \\
%&\le& \|\Pi_\mathcal{C}E\|_F + \|\Pi_\mathcal{C}P(\Omega)-P(\Omega)\|_F.  
%\end{eqnarray*}  
%Since $\Pi_\mathcal{C}(P(\Omega)-\Pi_\mathcal{L}P(\Omega))=0$, it follows that 
\begin{eqnarray}\label{eq:nnl ineq}
\|\Pi_\mathcal{C} A(\Omega)-P(\Omega)\|_F \le \|\Pi_\mathcal{C}E\|_F + \|\Pi_\mathcal{C}P(\Omega)-P(\Omega)\|_F.
\end{eqnarray} 
We now show that $\|\Pi_\mathcal{C}E\|_F$ is small due to condition \eqref{eq:correlated assumption}. 
Since $\mathcal{C}$ is a cone, whether $\|\Pi_\mathcal{C}E\|_F=0$ or $\|\Pi_\mathcal{C}E\|_F>0$, depending on the relative location of $E$ to $\mathcal{C}$. If $\|\Pi_\mathcal{C}E\|_F>0$ then
there exists $\nu\in\mathcal{C}$ with $\|\nu\|_F=1$ such that $\|\Pi_\mathcal{C}E\|_F=\langle\nu,E\rangle$. 
Let $\nu^{(r)}$ be the normalized predictor
$$\nu^{(r)} = \|\hat{P}^{(r)}(\Omega)\|_F^{-1}\hat{P}^{(r)}(\Omega)$$ 
and $\nu = \sum_{r=1}^m \lambda_r\nu^{(r)}$ for some $\lambda_1,...,\lambda_m\ge 0$. Then by \eqref{eq:correlated assumption}, 
\begin{eqnarray*}
1 = \|\nu\|_F^2 = \sum_{1\le r,s\le m}\lambda_r\lambda_s\langle\nu^{(r)},\nu^{(s)}\rangle \ge \delta\Big(\sum_{r=1}^m\lambda_r\Big)^2,
\end{eqnarray*} 
which implies $\sum_{r=1}^m \lambda_r \le \delta^{-1/2}$. Therefore
\begin{eqnarray*}
\|\Pi_\mathcal{C} E\|_F = \langle \nu,E\rangle = \sum_{r=1}^m \lambda_r \langle\nu^{(r)},E\rangle \le \delta^{-1/2}\max_{1\le r\le m} \langle\nu^{(r)},E\rangle.
\end{eqnarray*}
For each $\nu^{(r)}=\big(\nu^{(r)}_{ij}\big)$ with $\|\nu^{(r)}\|:=\max_{(i,j)\in\Omega}|\nu^{(r)}_{ij}|$, by Bernstein's inequality, 
\begin{eqnarray*}
\p\big(|\langle\nu^{(r)},E\rangle|>t\big)&\le& \exp\left(\frac{-t^2/2}{\sum_{(i,j)\in\Omega} \big(\nu_{ij}^{(r)}\big)^2 \text{var}(E_{ij})+\|\nu^{(r)}\|_\infty t/3}\right) \\
&\le& \exp\left(\frac{-t^2/2}{\|P(\Omega)\|_\infty+\|\nu^{(r)}\|_\infty t/3}\right).
\end{eqnarray*}      
Choosing $t \approx \max_{1\le r \le m}\|\nu^{(r)}\|_\infty\log(n+m) +(\|P(\Omega)\|_\infty\log(n+m))^{1/2}$ and applying the union bound, we obtain that with high probability,
\begin{eqnarray*}
\|\Pi_\mathcal{C}E\|_F \le C\delta^{-1/2}\left(\max_{1\le r \le m}\|\nu^{(r)}\|_\infty\log(n+m) +(\|P(\Omega)\|_\infty\log(n+m))^{1/2}\right).
\end{eqnarray*}
The proof is complete. 
%In our network setting we expect $\|\nu_i\|_\infty\approx 1/n$ and $\|\mu\|_\infty\approx d/n$, where $n$ is the number of nodes and $d$ is the average degree, so 
%\begin{eqnarray*}
%\|P_C\epsilon\| \approx (1-\delta)^{-1/2}\left(\frac{\log m}{n}+\sqrt{\frac{d\log m}{n}}\right)  \approx \sqrt{\frac{d\log m}{(1-\delta)n}}.
%\end{eqnarray*}
%From \eqref{eq:nnl ineq}, the error of the non-negative linear aggregation is bounded by
%\begin{eqnarray*}
%\|P_CY-P_L\mu\| \le \|P_CP_L\mu-P_L\mu\| + C\sqrt{\frac{d\log m}{(1-\delta)n}}.
%\end{eqnarray*}
%Note that $\|P_CP_L\mu-P_L\mu\|$ is the distance from the projection $P_L\mu$ of the true mean $\mu$ onto the linear subspace of $\theta_1,...,\theta_m$ to the cone $\mathcal{C}$, which is smaller than the smallest error $\min_{1\le i\le m}\|\mu-\theta_i\|$ of the estimates $\theta_i$. 
\end{proof}

\begin{proof}[Proof of Corollary~\ref{cor:comparison}]
Denote $E = A(\Omega)-P(\Omega)$. We can view $E$ and $\Pi_\mathcal{L}$ as a vector and a matrix in $\mathbb{R}^{|\Omega|}$ and $\mathbb{R}^{|\Omega|\times|\Omega|}$, respectively. 
Then $\|\Pi_\mathcal{L}A(\Omega)-\Pi_\mathcal{L}P(\Omega)\|_F^2 = \|\Pi_\mathcal{L}E\|_F^2
$. Denoting by $\circ$ the entry-wise product, we get
\begin{eqnarray}
\label{eq:expected norm}\e \|\Pi_\mathcal{L}E\|_F^2 &=& \e \langle\Pi_\mathcal{L}E,\Pi_\mathcal{L}E\rangle = \text{trace}(\Pi_\mathcal{L}\diag(P(\Omega)\circ (I-P(\Omega)))\Pi_\mathcal{L}^T)
\end{eqnarray} 
So we have
\begin{equation}\label{eq:relation1}
 m\sqrt{\rho} = m\cdot\min_{(i,j)\in\Omega} P_{ij}\cdot\Big(1-\max_{(i,j)\in\Omega} P_{ij}\Big)\le  \e \|\Pi_\mathcal{L}E\|_F^2 \le m\cdot\max_{(i,j)\in\Omega}P_{ij}.
 \end{equation}
Furthermore, by the Cauchy–Schwarz inequality, 
\begin{equation}\label{eq:relation2}
\left(\e \|\Pi_\mathcal{L}E\|_F\right)^2 \le \e \|\Pi_\mathcal{L}E\|_F^2\le m\cdot\|P(\Omega)\|_\infty.  
 \end{equation}

Since  $\|\Pi_\mathcal{L}\|=1$, by Hanson-Wright inequality \cite[Theorem 1.1]{Klochkov&Zhivotovskiy.hanson.wright.2020} and \eqref{eq:relation1}, for any $t\ge \max\{\e \|\Pi_\mathcal{L}E\|_F,1\}$ we have 
\begin{eqnarray*}
\mathbb{P}\left(\big|\|\Pi_\mathcal{L}E\|_F^2-\e \|\Pi_\mathcal{L}E\|_F^2\big|>t\right) &\le& \exp\left(-c\min\left\{\frac{t^2}{\left(\e \|\Pi_\mathcal{L}E\|_F\right)^2},t\right\}\right) \\
&\le& \exp\left(-c\min\left\{\frac{t^2}{ \e \|\Pi_\mathcal{L}E\|_F^2},t\right\}\right).
\end{eqnarray*}
Choosing $t=\left(m\log n\cdot\|P(\Omega)\|_\infty\right)^{1/2}$ and using \eqref{eq:relation2}, we see that
$$t > \max\{\e \|\Pi_\mathcal{L}E\|_F,1\}. $$
Then \eqref{eq:relation1} leads to 
\begin{equation}\label{eq:ols-upper}
\|\Pi_\mathcal{L}E\|_F^2 \le m\cdot\max_{(i,j)\in\Omega}P_{ij} + \left(m\log n\cdot\|P(\Omega)\|_\infty\right)^{1/2} 
\end{equation}
with high probability. Combining the assumption $m\ge 4\rho^{-1}\norm{P(\Omega)}_{\infty}\log n$ and \eqref{eq:relation1}, we can see that 
  $$\e \|\Pi_\mathcal{L}E\|_F^2 \ge m\sqrt{\rho} > 2\sqrt{m\norm{P(\Omega)}_{\infty}\log{n}} =2t > t$$   
  and therefore with high probability 
\begin{equation}\label{eq:ols-lower}
\|\Pi_\mathcal{L}E\|_F^2 \ge m\cdot\min_{(i,j)\in\Omega} P_{ij}\cdot\big(1-\|P(\Omega)\|_\infty\big) - \left(m\log n\cdot\|P(\Omega)\|_\infty\right)^{1/2} > \frac{m\sqrt{\rho}}{2}.
\end{equation}

Notice that \eqref{eq:ols-upper} directly indicates \eqref{eq:ols-oracle}. Now we have
\begin{eqnarray*}
\Delta &=& \norm{\hat{P}^{(\text{ols})}(\Omega) - P(\Omega)}_F^2 - \norm{\hat{P}^{(\text{nnl})}(\Omega) - P(\Omega)}_F^2\\
&=& \norm{\Pi_{\lcal}A(\Omega) - \Pi_{\lcal}P(\Omega)}_F^2 + \norm{\Pi_{\lcal}P(\Omega)-P(\Omega)}_F^2 \\
&& -\norm{\Pi_{\ccal}A(\Omega) - \Pi_{\lcal}P(\Omega)}_F^2- \norm{\Pi_{\lcal}P(\Omega) - P(\Omega)}_F^2\\
&\ge& \norm{\Pi_{\lcal}E}_F^2 - \norm{\Pi_{\ccal}E }_F^2 - \norm{\ \Pi_{\ccal}\Pi_{\lcal}P(\Omega)-\Pi_{\lcal}P(\Omega)}_F^2.
\end{eqnarray*}
On the other hand, by Theorem~\ref{thm:nnls} and \eqref{eq:ols-lower},
\begin{align*}
\Delta \ge  \frac{m\rho^{1/2}}{2} - \frac{C^2}{\delta}\epsilon^2- \|\Pi_\mathcal{C}\Pi_\mathcal{L} P(\Omega)-\Pi_\mathcal{L} P(\Omega)\|_F^2
\end{align*}
with high probability, and the proof is completed by using the basic inequality $(a+b)^2 \le 2(a^2+b^2)$ for $\epsilon^2$.
\end{proof}

\begin{proof}[Proof of Theorem~\ref{thm:self-regularizing}]
We will mainly follow the proof strategy of \cite{Slawski.NNLS.2011}. However, there is one key step in their proof (right after B.5) that may not go through. So our proof can be seen as a corrected version of that.

Recall the definition of $\hat{\beta}$ in \eqref{eq:nnls optimization}. Since this is a constrained linear regression problem, for the notation simplicity, let us denote $X_r = \hat{P}^{(r)}(\Omega)$, $Y=A(\Omega)$,  $\mu = P(\Omega)$, and $E =A(\Omega)-P(\Omega)= Y-\mu$. We view them as column vectors and further denote $X=(X_1,...,X_m)$. The optimization problem \eqref{eq:nnls optimization} is equivalent to
\begin{eqnarray}\label{eq:nnls linear form}
\hat{\pi} = \argmin_{\pi \succeq 0} \|Y-X\pi\|^2. 
\end{eqnarray}  
Consider the oracle parameter
$$\pi^*=\argmin_{\pi\succeq 0}\|\mu-X\pi\|^2$$
and define 
$\quad \delta=\pi^*-\pi$ for any $\pi \succeq 0$. In particular, we write $\hat{\delta}=\pi^*-\hat{\pi}.$
  
Our goal is to compare the NNL mixing estimate $X\hat{\pi}$ with the best linear approximation $X\pi^*$ in the noiseless setting when $\mu$ is known. We first rewrite the objective function in \eqref{eq:nnls linear form} as follows:  
\begin{eqnarray*}
\nonumber\|Y-X\pi\|^2 &=& \|\mu - X\pi^* + X(\pi^*-\pi) +E\|^2\\
&=& \|\mu - X\pi^*\|^2+\|X\delta\|^2 + \|E\|^2 +2E^T(\mu-X\pi^*) +2E^TX\delta.
\end{eqnarray*}
Note that the constraint $\pi\succeq 0$ is equivalent to $\delta\preceq\pi^*$, and only the second and the last terms of the expression above depend on $\delta$, it follows that
\begin{eqnarray}\label{eq:nnls delta}
\hat{\delta} = \argmin_{\delta\preceq \pi^*}\left\{\|X\delta\|^2+2E^TX\delta\right\}.
\end{eqnarray} 
Since the objective function on the right-hand side is zero when $\delta=0$, its value at the minimizer $\delta=\hat{\delta}$ is at most zero. Equivalently, we have   
\begin{equation*}
\|X\hat{\delta}\|^2 \le 2E^TX\hat{\delta}.
\end{equation*}
For a vector $x\in\mathbb{R}^m$, denote
$$S_+(x)=\{i\in[m]:x_i\ge 0\}, \quad S_{-}(x)=\{i\in[m]:x_i< 0\}.$$ 
Let $x_P$ be the vector obtained from $x$ by setting all entries within $S_{-}(x)$ to zero and $x_N=x-x_P$. The inequality above implies 
\begin{equation}\label{eq:Xdelta bound}
\|X\hat{\delta}\|^2 \le2\|E^TX\|_\infty\|\hat{\delta}\|_1 \le 2\|E^TX\|_\infty\big(\|\hat{\delta}_P\|_1+\|\hat{\delta}_N\|_1\big) \le 2\|E^TX\|_\infty\big(\|\pi^*\|_1+\|\hat{\delta}_N\|_1\big),
\end{equation}
where $\|\hat{\delta}_P\|_1\le \|\pi^*\|_1$ because $\hat{\delta}=\pi^*-\hat{\pi}\le \pi^*$ for $\hat{\pi}\succeq 0$. It remains to bound $\|\hat{\delta}_N\|_1$. 

Denote $\Sigma=X^TX$. For each $\delta\le \pi^*$, let $\Sigma_{PP}$, $\Sigma_{NN}$ and $\Sigma_{NP}$ be the matrices obtained from $\Sigma$ by setting all entries with indices $(i,j)$ outside $S_+(\delta)\times S_+(\delta)$, $S_{-}(\delta)\times S_{-}(\delta)$ and $S_{-}(\delta)\times S_{+}(\delta)$ to zero, respectively.   
Then \eqref{eq:nnls delta} is equivalent to
\begin{eqnarray*}
\hat{\delta}=\argmin_{\delta\preceq \pi^*}\left\{\delta_P^T\Sigma_{PP}\delta_P+\delta_N^T\Sigma_{NN}\delta_N+2\delta_N^T\Sigma_{NP}\delta_P+2E^TX(\delta_N+\delta_P)\right\}.
\end{eqnarray*}

Define $\tcal = \{x \in \bR^m: x\le\pi^*, x_i \le 0, i \in S_-(\hat{\delta}),  x_i=0, i\in S_+(\hat{\delta})\}$. Replacing $\delta_P$ in the objective function above on the right-hand side with $\hat{\delta}_P$, we see that 
\begin{eqnarray*}
\hat{\delta}_N = \argmin_{\delta_N\in\tcal}\left\{\delta_N^T\Sigma_{NN}\delta_N+2\delta_N^T\Sigma_{NP}\hat{\delta}_P+2E^TX\delta_N\right\}.
\end{eqnarray*}
%where $\tcal$ is the set of all vectors $x\le\beta^*$ such that $x_i\le 0$ for all $i\in S_-(\hat{\delta})$ and $x_i=0$ for all $i\in S_+(\hat{\delta})$. 
Since the objective function on the right-hand side is zero when $\delta_N=0\in\Omega$, 
\begin{eqnarray*}
\hat{\delta}_N^T\Sigma_{NN}\hat{\delta}_N+2\hat{\delta}_N^T\Sigma_{NP}\hat{\delta}_P+2E^TX\hat{\delta}_N \le 0.
\end{eqnarray*}
Denote $\|\Sigma\|_\infty=\max_{1\le r,s\le m}|\Sigma_{rs}|$. By the self-regularizing property \eqref{eq:self-reg}, we have
\begin{eqnarray*}
0 &\ge& \kappa\|\hat{\delta}_N\|_1^2\|\Sigma\|_\infty -2\|\hat{\delta}_N\|_1\|\hat{\delta}_P\|_1\|\Sigma\|_\infty+2E^TX\hat{\delta}_N \\
&\ge& \kappa\|\hat{\delta}_N\|_1^2\|\Sigma\|_\infty -2\|\hat{\delta}_N\|_1\|\hat{\delta}_P\|_1\|\Sigma\|_\infty-2\|E^TX\|_\infty\|\hat{\delta}_N\|_1.
\end{eqnarray*} 
It follows that
\begin{eqnarray}\label{eq:delta-N-bound}
\|\hat{\delta}_N\|_1 \le \frac{2}{\kappa}\left(\|\hat{\delta}_P\|_1+\frac{\|E^TX\|_\infty}{\|\Sigma\|_\infty}\right)\le \frac{2}{\kappa}\left(\|\pi^*\|_1+\frac{\|E^TX\|_\infty}{\|\Sigma\|_\infty}\right).
\end{eqnarray}
From \eqref{eq:Xdelta bound} and \eqref{eq:delta-N-bound}, we get
\begin{eqnarray*}
\|X\hat{\pi}-X\pi^*\|^2 \le \frac{4\|E^TX\|_{\infty}^2}{\kappa \|\Sigma\|_\infty}+\frac{(2\kappa+4)\|E^TX\|_\infty\|\pi^*\|_1}{\kappa}.
\end{eqnarray*}
By Bernstein's inequality on the Bernoulli error $E$, for each $1\le r\le m$, we have
\begin{eqnarray*}
\mathbb{P}\left(|E^TX_r|>t\right) \le 2\exp\left(\frac{-t^2/2}{\|P(\Omega)\|_\infty\|\hat{P}^{(r)}(\Omega)\|_F^2+t\|\hat{P}^{(r)}(\Omega)\|_\infty/3}\right).
\end{eqnarray*}
Choosing $t\approx \|\hat{P}^{(r)}(\Omega)\|_\infty\log(n+m) + \|\hat{P}^{(r)}(\Omega)\|_F(\|P(\Omega)\|_\infty\log(n+m))^{1/2}$
for each $r$ and using the union bound across $r$, we obtain the final bound for $\|X\hat{\pi}-X\pi^*\|^2 = \norm{\hat{P}(\Omega) - \Pi_{\ccal}P(\Omega)}_F^2$ with
$$\Phi = \max_r\|\hat{P}^{(r)}(\Omega)\|_\infty + \|P(\Omega)\|_\infty^{1/2}\max_r\|\hat{P}^{(r)}(\Omega)\|_F = \max_r\|\hat{P}^{(r)}(\Omega)\|_\infty + \|P(\Omega)\|_\infty^{1/2}\norm{\Sigma}_{\infty}.$$
The proof is complete. 
\end{proof}